\pdfoutput=1

\documentclass[11pt]{article}

\usepackage[final]{acl}

\usepackage{times}
\usepackage{latexsym}
\usepackage{tipa}
\usepackage{amsmath}
\usepackage{booktabs}
\usepackage{multirow}
\usepackage{graphicx} 
\usepackage{longtable}
\usepackage[T1]{fontenc}
\usepackage[utf8]{inputenc}

\usepackage[utf8]{inputenc}

\usepackage{microtype}

\usepackage{inconsolata}

\usepackage{graphicx}
\usepackage{tipa}
\title{ZIPA: A family of efficient models for multilingual phone recognition}

\author{Jian Zhu \\
  University of British Columbia \\
  \texttt{jian.zhu@ubc.ca} \\\And
  Farhan Samir \\
  University of British Columbia \\
  \texttt{fsamir@mail.ubc.ca} \\\AND
  Eleanor Chodroff \\
  University of Zurich \\
  \texttt{eleanor.chodroff@uzh.ch} \\\And
  David R. Mortensen \\
  Carnegie Mellon University \\
  \texttt{dmortens@cs.cmu.edu}
  }

\begin{document}
\maketitle
\begin{abstract}
We present \textsc{Zipa}, a family of efficient speech models that advances the state-of-the-art performance of crosslinguistic phone recognition. We first curated \textsc{IpaPack++}, a large-scale multilingual speech corpus with 17,132 hours of normalized phone transcriptions and a novel evaluation set capturing unseen languages and sociophonetic variation. With the large-scale training data, \textsc{Zipa}, including transducer (\textsc{Zipa-T}) and CTC-based (\textsc{Zipa-Cr}) variants, leverage the efficient Zipformer backbones and outperform existing phone recognition systems with much fewer parameters. Further scaling via noisy student training on 11,000 hours of pseudo-labeled multilingual data yields further improvement. While \textsc{Zipa} achieves strong performance on benchmarks, error analysis reveals persistent limitations in modeling sociophonetic diversity, underscoring challenges for future research. 
\end{abstract}

\section{Introduction}
The International Phonetic Alphabet (IPA) provides a theoretically unified discrete representation of all known human speech sounds \cite{international1999handbook}. IPA transcriptions capture major articulatory contrasts in speech sounds, including the voicing status, place of articulation, manner of articulation, and tongue positions \cite{ladefoged2014course}. In phonetics, the IPA is the major tool to document speech sounds across the world's languages, thanks to its universality. Therefore, developing speech technology that can transcribe multilingual speech into phones, or IPA symbols can significantly facilitate language documentation, especially for low-resource languages. 

Even beyond linguistics, phone transcriptions are also widely used in various speech technologies, including multilingual pretraining \cite[e.g.,][]{feng23b_interspeech,yusuyin2025whistle}, speech synthesis \cite[e.g.,][]{liu2023pronunciation}, speech enhancement \cite[e.g.,][]{liu2021phoneme,pirklbauer2023evaluation}, pronunciation assessments \cite[e.g.,][]{speechocean762,gong2022transformer}, and voice conversion \cite[e.g.,][]{lee2022duration,shan2024phoneme}. 

In this study, we present state-of-the-art phone recognition systems that can transcribe speech into IPA symbols crosslinguistically. Our core contributions are summarized as follows. 
\vspace{-0.05in}
\begin{itemize}
    \setlength{\parskip}{1pt}  
    \item First, we curate \textsc{IpaPack++}, a 17,132-hour open-source speech corpora with G2P-generated phonetic transcriptions. We also design an evaluation set containing rich crosslinguistic and sociophonetic variation. 
    \item Second, we present a series of state-of-the-art phone recognition models, the transducer \textsc{Zipa-T} and the CTC-based \textsc{Zipa-Cr} in two sizes (64M and 300M). Trained on the \textsc{IpaPack++}, even the 64M \textsc{Zipa} models outperform previous phone recognition models with 300M parameters, while being more computationally efficient. 
    \item Third, we further applied noisy student training on \textsc{Zipa-Cr} models with 11k hours of pseudo-labeled speech in more than 4,000 languages, resulting in state-of-the-art performance on phone recognition. 
    \item Finally, we conducted error analyses on the model prediction, showing that current phone recognition models, despite the impressive performance, are still struggling with predicting sociophonetic variation. Our analysis thus reveals a critical, overlooked limitation of current data curation practices in training universal phone recognition models.
\end{itemize}
We will release all training and evaluation data, pre-trained models, and the code under permissive licenses at \url{https://github.com/lingjzhu/zipa}.

\section{Background}
\subsection{Multilingual phone recognition}
Early efforts in automatic speech recognition in the 1970s were centered on prediction of phones \citep{li2017divination}. 
There has been a resurgence in interest in phonetic transcription \citep{li2020universal, gao2021zero, xu22b_interspeech, taguchi23_interspeech, glocker23_interspeech, samir2024efficiently}. These models have proven indispensable for transcribing speech in oral languages \citep{lane2021local}, and have high potential for facilitating cross-linguistic phonetic analysis \citep{chodroff-etal-2024-phonetic}. 
Most systems are trained through fine-tuning pretrained multilingual models like XLS-R and Whisper \citep{babu2021xls,radford2023robust} on large audio-transcript archives like VoxClamantis \citep[e.g.,][]{salesky-etal-2020-corpus} or X-IPAPack \citep{samir2024efficiently}. But the transcripts are semi-automatically generated through applying G2P models to orthographic transcripts.  

Still, there remain significant challenges with training reliable phonetic transcript models for the world's languages. First, the linguistic diversity of the datasets needs to be considerable in order to transcribe audio from any language. As shown in \citet{samir2024efficiently}, collecting reliably transcribed audio-transcript pairs is far from trivial, as algorithmic curation pipelines for obtaining massively multilingual transcribed audio archives can fail. Importantly, these failures manifest when the G2P model is not calibrated for the language variety represented by the audio. To this end, we collect the \textsc{IpaPack++} dataset (Section~\ref{sec:data}), comprising 17K+ hours of reliable phonetically transcribed audio in 88 languages.

Moreover, another potential challenge is that G2P models tend to capture dictionary-like pronunciations for the standard dialect of the language, thereby failing to capture pronunciation patterns in audio for different sociolects. Therefore, we specifically design evaluation datasets rich in sociophonetic variation to evaluate whether the phone recognition models are simply memorizing the standard pronunciations. 



\subsection{Phone recognition is subjective}

While the IPA provides a universal representation of speech sounds, applying IPA crosslinguistically still poses many challenges. The acoustic-phonetic details of a given speech segment can vary considerably across speakers and languages. For example, voice onset time (VOT) is commonly known as the primary acoustic correlate for separating voiceless from voiced stops across languages \cite{abramson2017voice}. However, the absolute values of VOT vary substantially across languages \cite{cho1999variation, chodroff2019covariation}, which cannot be easily captured via discrete IPA symbols. 

Therefore, phonetic transcription remains a highly subjective process, affected by the linguistic backgrounds or theoretical orientations of the transcriber. In transcription practices, strict transcriptions are not always necessary or achievable because many non-contrastive phonetic details are usually irrelevant in a given analysis linguistic analysis \cite{anderson2023variation,kerswill1990validity,shriberg1991reliability}. \citet{shriberg1991reliability} conducted a meticulous comparison of broad and narrow transcriptions by trained personnel. For broad transcriptions, the agreement between human annotators was generally acceptable. However, for narrow transcriptions involving diacritics, the agreements were ``\textit{below acceptable reliability boundary levels, even at the least strict agreement criteria}" \cite{shriberg1991reliability}.

Given the subjectivity of phone transcriptions, we focus our efforts on \textbf{broad transcription}. Broad transcription encodes only the most salient phonetic features, usually the base vowels and consonants with infrequent use of diacritics.  This is in contrast to \textbf{narrow transcription}, where the transcriber will try to transcribe as many subphonemic or phonetic details as possible with the frequent use of diacritics \cite{ladefoged2014course}. Since objectively true transcriptions might not exist, we evaluate our transcriptions with phonetic feature error rates (PFER) \cite{taguchi23_interspeech}, measuring the distance between binary articulatory features.


\begin{figure*}
    \centering
    \includegraphics[width=\linewidth]{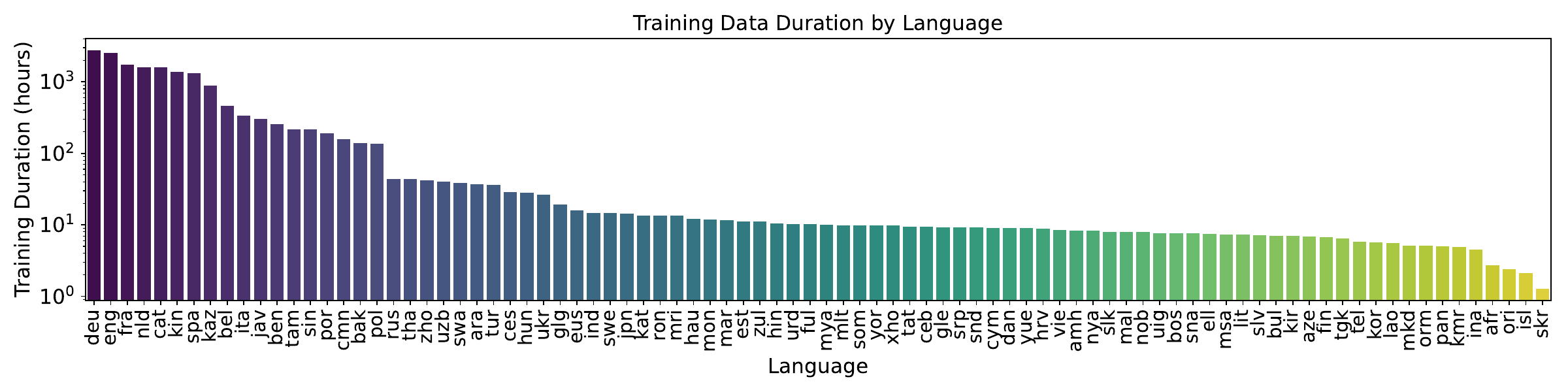}
    \caption{The distribution of labeled training data duration by language, totaling 17,132 hours.}
    \label{fig:dist}
\end{figure*}

\section{Data} \label{sec:data}
First, we have created \textsc{IpaPack++}, one of the largest phone-based speech corpora in 88 languages, totaling 17,132 hours. 
While the original \textsc{Ipa Pack} \cite{zhu-etal-2024-taste} provides 2000+ hours of speech in 100+ languages, upon careful inspection, we noticed several shortcomings. First, the IPA transcriptions were not normalized across the corpus, such that different Unicode encodings were present for the same phone. Some non-IPA Unicode symbols were also present due to artifacts in preprocessing. Second, the original dataset was more suitable for keyword spotting than ASR as half of the corpus was short clips of words taken from continuous recordings. 

\subsection{Data selection}
To address some of these limitations and expand our efforts, we have created a large-scale speech dataset for phone recognition. The datasets are recreated from \textsc{Ipa Pack} \cite{zhu-etal-2024-taste}, Common Voice 16.0 \cite{ardila-etal-2020-common}, LibriSpeech \cite{panayotov2015librispeech}, Multilingual LibriSpeech \cite{pratap20_interspeech}, Aishell-1 \cite{bu2017aishell}, crowd-sourced speech corpora for Javanese, Sinhala, and Bengali \cite{kjartansson-etal-sltu2018}, IISc-MILE Tamil ASR Corpus \cite{mile_1,mile_2}, Kazakh Speech Dataset \cite{mansurova-kadyrbek-2023-kazakh-speech-dataset} and Kazakh Speech Corpus \cite{khassanov-etal-2021-crowdsourced}. CharsiuG2P \cite{zhu22_interspeech} and Epitran \cite{mortensen-etal-2018-epitran} were used to automatically create phonemic transcriptions of available languages. 

After preprocessing, we ended up with around 17,135 hours of training data with G2P-generated transcriptions in 88 languages. The language distribution of the \textsc{IpaPack++} is shown at Figure~\ref{fig:dist}. A complete breakdown of individual languages can be found in Appendix~\ref{app:dataset}.

\begin{table*}[]
\small
\centering
\resizebox{\textwidth}{!}{%
\begin{tabular}{lrl}
\toprule
Dataset    & Dur. & Description \\\midrule
\texttt{Doreco}     & 19 hrs      & 45 languages collected and transcribed by field linguists.     \\
\texttt{VoxAngeles} & 1.5 hrs     & A set of individual word recordings from 95 languages.  \\\midrule
\texttt{Buckeye}    & 8 hrs & A collection of sociolinguistic recordings, carefully annotated by trained phoneticians. \\
\texttt{L2-Standard} & 4 hrs & L2-ARCTIC speech corpus with dictionary-based phonetic transcriptions. \\
\texttt{L2-Perceived} & 4 hrs & L2-ARCTIC speech corpus with human transcriptions of the actual pronunciation. \\\midrule
\texttt{Seen languages} & 65 hrs & Test sets from Aishell, LibriSpeech, and the Multilingual LibriSpeech (except for English). \\
\bottomrule
\end{tabular}%
}
\caption{A list of the evaluation datasets. These datasets cover a wide range of languages and sociophonetic conditions. }
\label{tab: eval_dataset}
\end{table*}

\subsection{Tokenization} 
The prior state-of-the-art universal phone recognizer \cite{xu22b_interspeech} adopted a data-driven approach to tokenization. However, this approach is not without problems. The phone tokenizer includes plain phones as well as phone combinations that are highly language-specific. For example, it uses a numerical representation of Mandarin tones rather than the standard IPA tone notations, also known as Chao tone letters. The numerical representation represents symbolic phonological contrasts of the tones, whereas the Chao tone letters reflect aspects of the phonetic realization like the f0 contour. Overall, though, using inconsistent symbols can limit knowledge sharing across languages \cite{zhu-etal-2024-taste}. 

We further made a systematic effort to normalize IPA encodings. In the first round of filtering, PHOIBLE \cite{moran2014phoible} was used as a reference to determine whether a phone was legitimate. Illegitimate phones were corrected: 1) phones with more than 3 diacritics can be overly complex to transcribe, so they are simplified to no more than one; 2) phones with inconsistent Unicode encodings, such as [g] (Unicode: U+0067) and [\textipa{g}] (Unicode: U+0261), are unified in one representation. Since we only focused on broad transcriptions, our final tokenizer only consists of all individual IPA symbols and the 15 most frequent diacritics from the IPA chart. Each diacritic is encoded as a separate token to reduce the vocabulary size.

\subsection{Evaluation set}
\paragraph{Evaluating on seen languages}
We used the test set of several publicly available datasets to evaluate model performance. The G2P-generated phone transcriptions are quite noisy \citep{samir2024efficiently}, especially for low-resource languages. Therefore, we selected the test sets from Aishell-1 \cite{bu2017aishell}, Librispeech \cite{panayotov2015librispeech} and Multilingual LibriSpeech (MLS) \cite{pratap20_interspeech}, where the phone transcription quality was determined to be good upon our inspection. 

\paragraph{Evaluating on unseen languages}
In order to test how universal phone recognition models generalize across languages, we reserved the VoxAngeles \cite{chodroff-etal-2024-phonetic}, a clean version of the UCLA Phonetic Corpus \cite{li2021multilingual}, and DoReCo \cite{paschen2020building} for evaluation on unseen languages. Both datasets consist of speech recordings collected from fieldwork and transcribed phonetically by trained linguists. 

\paragraph{Evaluating on sociophonetic variation}
Most phone recognition models are trained and evaluated on dictionary pronunciations generated from pronunciation dictionaries and G2P models. These training and evaluation data might not reflect the actual pronunciation in spontaneous speech. We also measure how phone recognition models can predict actual phonetic variation. Such evaluation can serve to assess whether phone recognition models are suitable for tasks like pronunciation assessment and sociophonetic transcriptions. 

Here we utilize L2 ARCTIC \cite{zhao2018l2} and the Buckeye Corpus \cite{pitt2005buckeye}, both of which contain highly variable English speech carefully transcribed by professional linguists. For the Buckeye Corpus, we segmented all recording files into individual utterances between 20 to 50 phonemes, delimited by silent intervals ($\geq 200$ms) \cite{fuchs22_interspeech}. For L2 ARCTIC, we used the original segmentation but generated two versions of transcriptions, one for dictionary pronunciations of the prompts and one for the perceived pronunciations annotated by linguists.

\section{Method}
Some prior studies in universal phone recognition  leverage knowledge of the language's phonemic inventory \cite{li2020universal,glocker23_interspeech}. However, the inventory is a static, abstract description of the phonological system of a language, only capturing a limited, idealized variation of speech. Many speech variations within a language can go beyond the inventory. In many applications of phone recognition such as pathological speech assessment, pronunciation assessment, and sociophonetics, transcribing speech into phones as it is actually articulated is important. Therefore, in our proposed models, we did not directly incorporate language-specific inventory knowledge, noting that such knowledge can also be incorporated in post-processing \cite{xu22b_interspeech}.

\subsection{Zipformer}
Pretrained self-supervised models such as XLS-R \cite{babu22_interspeech} and Whisper \cite{radford2023robust} have been utilized as base models for fine-tuning in prior studies \cite{xu22b_interspeech,taguchi23_interspeech,glocker23_interspeech,samir2024efficiently}.
However, fine-tuning these transformer models on our large-scale dataset is prohibitively expensive with an academic computing budget. For example, Whisper pads every input utterance, regardless of their lengths, to chunks of 30 seconds,  allocating many computations to padding tokens that do not contribute to inference. Moreover, its autoregressive decoding is also highly inefficient.

Instead, we adopt Zipformer \citep{yao2023zipformer}, a transformer encoder model with U-Net style downsampling and upsampling layers \cite{ronneberger2015u} as the base architecture. Compared to the vanilla transformers (e.g., Wav2Vec2 and XLS-R), Conformer \cite{gulati20_interspeech}, Branchformer \cite{peng2022branchformer} and E-Branchformer \cite{kim2023branchformer}, Zipformer has demonstrated superior ASR performance with less compute \citep{yao2023zipformer}. Zipformer achieves such compute efficiency through reusing attention weights across layers, and progressively downsampling speech in the middle layers and upsampling to the output resolution in later layers. 

\subsection{CR-CTC}
We use the Connectionist Temporal Classification (CTC) loss \cite{graves2006connectionist} because it enables efficiently parallelized predictions and has maintained competitive results compared to an encoder-decoder architecture \cite{peng2024owsm}. Specifically, we adopted Consistency-Regularized CTC \cite{yao2025crctc} for our phone recognition model. 

Given a speech-transcription pair $(\mathbf{x},\mathbf{y})$, we fit an ASR model $f(\cdot)$. For the input speech spectrogram $\mathbf{x}$, $\mathbf{x}^{(a)}$ and $\mathbf{x}^{(b)}$ are two different augmented views generated through SpecAugment \cite{park19e_interspeech}. Two CTC output frame-wise distributions are generated through $\mathbf{z}^{(a)}=f(\mathbf{x}^{(a)})$ and $\mathbf{z}^{(b)}=f(\mathbf{x}^{(b)})$. 
Then the CR-CTC loss is formulated as:
\begin{align*}
\mathcal{L}_{CR-CTC} &= \frac{1}{2}\left( \mathcal{L}_{CTC}(\mathbf{z}^{(a)}, \mathbf{y}) + \mathcal{L}_{CTC}(\mathbf{z}^{(b)}, \mathbf{y}) \right) \\
            &\quad + \alpha \mathcal{L}_{CR}(\mathbf{z}^{(a)}, \mathbf{z}^{(b)})
\end{align*}
In addition to the regular CTC loss $\mathcal{L}_{CTC}$, $\mathcal{L}_{CR}$ is used to regularize the output distributions with Kullback-Leibler (KL) divergence between two frame-wise distributions at the same time step. The CR loss is defined as:
\begin{align*}
\mathcal{L}_{CR}(\mathbf{z}^{(a)}, \mathbf{z}^{(b)}) &= \frac{1}{2}\sum_{t=1}^{T}D_{KL}\left(sg(z_t^{(b)}),z_t^{(a)}\right) \\
            &\quad + D_{KL}\left(sg(z_t^{(a)}),z_t^{(b)}\right)
\end{align*}
where $sg(\cdot)$ is the stop-gradient operation and $D_{KL}$ is the KL divergence. $\mathcal{L}_{CR}$ performs self-distillation between outputs of two different augmented input views, mitigating overfitting. It has been shown to outperform regular CTC loss and RNN-T loss \cite{yao2025crctc}. 

We used the original CR-CTC implementation \cite{yao2025crctc} with minor modifications. The output temporal resolution is 25 Hz in the original Zipformer model. Yet this resolution is too short for phone sequences, which are significantly longer than text tokens. We upsampled the output resolution to 50 Hz to present numerical errors when computing the CTC loss. We also trained two variants of CR-CTC models: \textsc{Zipa-Cr-small} with 64M parameters and \textsc{Zipa-Cr-large} with 300M parameters.

\subsection{Transducer}
We also trained Zipformer-based transducer models \citep{yao2023zipformer}. The original RNN-T loss for tranducers is computation- and memory-intensive, so we utilized the memory-efficient pruned RNN-T loss \cite{kuang22_interspeech}. In the transducer, we used Zipfomer as the encoder and the stateless decoder with 1D convolutional layers \cite{ghodsi2020rnn}. We trained two variants with non-causual attention: \textsc{Zipa-t-small} with 65M parameters and \textsc{Zipa-t-large} with 302M parameters. 

\subsection{Noisy student training}
Prior studies have shown that noisy student training \cite{park20d_interspeech}, or training on pseudo-labels can reliably improve multilingual ASR performance \cite{hwang2022large,hwang22c_interspeech,ramirez2024anatomy}. We generated phone pseudo-labels for two unannotated multilingual speech datasets, VoxLingua-107 and MMS ulab v2. VoxLingua-107 \cite{valk2021voxlingua107} consists of speech recordings without transcriptions from 107 languages, totaling 6,628 hours. MMS ulab v2 \cite{chen-etal-2024-towards-robust} is a 6,700-hour speech dataset in 4,023 languages, a reproduction of the original dataset for training Meta MMS \cite{pratap2024scaling}. As our labelled training data only included 88 unique languages, these two datasets can tremendously enrich the language diversity of our training data. 

We used all four Zipformer-based phone recognition models to generate the pseudo-labels for these two multilingual corpora, and computed the pairwise Phonetic Feature Error Rate (PFER) with PanPhon \cite{mortensen-etal-2016-panphon}. The consistencies of predictions between models were used as a heuristic to filter out bad predictions. Speech samples with an averaged pairwise PFER higher than the 80 percentile were ultimately excluded. We used pseudo-labels from \textsc{Zipa-Cr-large} as the final transcriptions for simplicity. Finally, we obtained pseudo-labels for 11,851 hours of multilingual speech in around 4,000 languages. We continued to train the CR-CTC models by mixing both the original dataset and pseudo-labelled dataset. The loss function was formulated as below.
$$
\mathcal{L}_{mixed} = \mathcal{L}_{CR-CTC} + \lambda\cdot\mathcal{L}_{CR-CTC}^{Pseudo}
$$
The hyperparameter $\lambda$ was set to 0.5 to downscale the weights of the noisy pseudo-labels. We adopted noisy student training to train \textsc{Zipa-Cr-Ns-small} and \textsc{Zipa-Cr-Ns-large}, both of which were initialized from pretrained checkpoints of \textsc{Zipa-Cr-small} and \textsc{Zipa-Cr-large} respectively. 

\paragraph{No-Diacritic Models} Our error analysis suggested that many recognition errors were associated with diacritics. During noisy student training, we also trained two variants of  \textsc{Zipa-Cr-small} and \textsc{Zipa-Cr-large} without diacritics. We maintained the exact same training settings, but removed all diacritics from all training data. For consistency, these models were also evaluated with the same evaluation data but without diacritics.  

\begin{table*}[]
\centering
\small
\resizebox{\textwidth}{!}{%
\begin{tabular}{rrcrrrrrrrrrr}
\toprule
Model & Param. & Iters & \texttt{eng-c} & \texttt{eng-o} & \texttt{ger} & \texttt{por} & \texttt{fre} & \texttt{spa} & \texttt{dut} & \texttt{ita} & \texttt{cmn} & \texttt{Avg.}\\\midrule
\texttt{Allosaurus}      &     11M       &      -     &    4.18       &  6.21   &  30.26  &  33.09 &  32.77    &  28.02  & 33.29 & 26.57 & 6.64  & 22.33 \\
\texttt{W2V2P-lv-60-ft} & 300M & - & 4.09 & 4.26 & 18.11 & 23.47 &  28.63   &  7.97 & 27.27  & 8.63 &  6.82 & 14.36\\
\texttt{W2V2P-xlsr-53-ft}       &     300M       &    -       &    5.45       &     5.35   &   11.61  &  18.80   &  26.59   &  5.14 & 20.91 & 6.93 & 6.20 & 11.88  \\
\texttt{MultIPA}$^*$      &   300M         &    -       &   11.26        &  10.86   &  27.02   &  25.05   &   31.31  &  12.02   & 32.15 & 11.22 & 8.35 & 18.80 \\
\texttt{WhisperPPT} & 244M & - & 6.36 & 7.39 & 20.40 & 18.29 & 26.85 & 6.89 & 13.29 &  5.52 & 2.03 & 11.89 \\\midrule
\textsc{Zipa-T-small}     &    65M        &    300k               & 1.17    &  2.14   &  4.66   &  18.20   &  16.68  &  2.34 & 6.84 & 5.05 & 0.72 & 6.42  \\
\textsc{Zipa-T-large}     &     302M       &   300k        &     0.70     &  1.35   &  3.76   &  6.52   &  5.32   &   1.85  &  5.33 & 8.91 & 0.52 & 3.80 \\
\textsc{Zipa-T-small}     &    65M        &    500k       &      0.95     &  1.67   & 3.51    &   17.01  &  7.49   & 2.08 &  5.49   & 2.66 & 0.78  & 4.62  \\
\textsc{Zipa-T-large}     &     302M       &    500k       &       \textbf{0.61}   &  \textbf{1.19}   &   3.38  &   \underline{5.96}  &  \textbf{4.52}   &  \textbf{1.69}   & \textbf{4.62}  & \textbf{1.91} & \underline{0.44} & \textbf{2.70}  \\\midrule
\textsc{Zipa-Cr-small}     &     64M       &    300k      & 2.36    &  3.29   &  14.11   &  20.19   &  18.19   &  4.07 & 9.69 & 8.27 & 1.59  & 9.08  \\
\textsc{Zipa-Cr-large}     &     300M       &    300k       &     1.07      &   1.92  &   3.70  &  21.15   &  5.47   & 2.37    &   5.25 &  2.28 & 0.55 & 4.86 \\
\textsc{Zipa-Cr-small}     &     64M       &     500k      &      1.15     &  2.23   &  3.56   &  18.19   & 6.13    &  2.74   &  7.27 & 8.47& 0.84 & 5.62  \\
\textsc{Zipa-Cr-large}     &     300M       &    500k       &      0.77     &  1.49   & \underline{3.34}    &  7.10   &  4.99   &  2.58   & 5.23  & 2.23& 0.54 & 3.14 \\\midrule
\textsc{Zipa-Cr-Ns-small}     &     64M       &     700k      &  0.75 &  1.51 &  3.41 & 8.56 & 4.87 & 2.36 &  4.9 & \underline{2.19} & 0.50 & 3.22 \\
\textsc{Zipa-Cr-Ns-large}     &     300M       &    800k       &  \underline{0.66} & \underline{1.29} & \textbf{3.07} & \textbf{5.47} &  \underline{4.53} & \underline{1.98} & \underline{4.86} & 2.23 & \textbf{0.38}  & \underline{2.71}   \\\midrule
\textit{No diacritics$^{**}$} \\\midrule
\textsc{Zipa-Cr-Ns-small}     &     64M       &     700k      &  0.78 &  1.51 &  3.25 & 8.70 & 4.83 & 2.31 &  4.73 & 2.10 & 0.50 & 3.02 \\
\textsc{Zipa-Cr-Ns-large}     &     300M       &    780k       &  0.65 & 1.28 & 2.95 & 4.92 &  4.55 & 2.24 & 4.68 & 2.20 & 0.41  & 2.65   \\
\bottomrule
\end{tabular}
}
\caption{Main PFER results on seen languages. $^*$Some languages were not seen by \texttt{MultiIPA}. $^{**}$Diacritics were removed for both training and evaluation sets, so results are not directly comparable with other models. \textbf{Notations:} \textsc{T} - Transducer; \textsc{Cr} - Consistency-regularized CTC; \textsc{Ns} - Noisy student training. }
\label{tab:seen}
\end{table*}

\section{Experiments}

\subsection{Implementation} Our experiments were structured within the Next-gen Kaldi framework. We used \texttt{lhotse}\footnote{\url{https://github.com/lhotse-speech/lhotse}} to manage data loading and augmentation, \texttt{icefall}\footnote{\url{https://github.com/k2-fsa/icefall}} for training and evaluation, and \texttt{k2}\footnote{\url{https://github.com/k2-fsa/k2}} for the pruned transducer loss. The inputs to all models are the 80-dimensional Mel Frequency Cepstral Coefficients (MFCCs).
We used the Scaled Adam optimizer, which was shown to work better with Zipformer than Adam \citep{yao2023zipformer}. All models were trained from scratch with randomly initialized weights. During evaluation, the final model for each variant was the averaged model from the last 10 checkpoints. Simple greedy decoding was used to generate predictions in all conditions. We trained all small models with an A40 40G GPU and all large models with 2 A100 40G GPUs. 
Detailed hyperparamters are described in Appendix~\ref{app:training}.

\begin{table*}[]
\small
\centering
\resizebox{\textwidth}{!}{%
\begin{tabular}{@{\extracolsep{4pt}}rcrrrrrrr}
\toprule
\multirow{2}{*}{Model} & \multirow{2}{*}{Param.} & \multirow{2}{*}{Iters}  & \multicolumn{2}{c}{\bf Unseen} & \multicolumn{3}{c}{\bf Sociophonetic} & \multirow{2}{*}{Avg.} \\\cline{4-5}\cline{6-8}
                       &                             &                        & \texttt{Doreco}     & \texttt{VoxAngeles}    & \texttt{L2-Standard}  & \texttt{L2-Perceived}      & \texttt{Buckeye}       \\\midrule
\texttt{Allosaurus}      &     11M       &    -       &  8.89         &   1.35 &  5.21 & 5.72 & 5.04 & 5.24 \\
\texttt{W2V2P-lv-60-ft} & 300M & - & 6.13 & 0.66 & 2.89 & 3.95 & \textbf{3.85} & 3.49 \\
\texttt{W2V2P-xlsr-53-ft}      &     300M       &    -       &  \underline{5.94}    &  \textbf{0.58}    &  3.69 & 4.09   &      3.97 & 3.65 \\
\texttt{MultIPA}      &   300M         &     -      &    6.55       &  \underline{0.62}    & 5.86 & 5.88 & 5.94 & 4.97   \\
\texttt{WhisperPPT} & 244M & - & 9.31 & 0.91 & 6.44 & 6.65 & 6.80 & 6.02 \\\midrule
\textsc{Zipa-T-small}     &    65M        &   300k               &  7.20 & 0.71 & 2.26 & 3.85 & 4.00 & 3.60     \\
\textsc{Zipa-T-large}     &     302M       &     300k      &       7.27     &   0.88   & 1.79 & 3.67 & 3.95 & 3.51\\
\textsc{Zipa-T-small}     &    65M        &     500k      &       8.72    &    0.78  &  1.99 & 3.80 & 3.97 & 3.85  \\
\textsc{Zipa-T-large}     &     302M       &    500k       &        8.05    &  0.88   &  \textbf{1.68} & \textbf{3.63} & 3.94 & 3.63 \\\midrule
\textsc{Zipa-Cr-small}     &     64M       &    300k        &  5.97 & 0.78 & 3.42 & 5.13 & 4.58  & 3.97    \\
\textsc{Zipa-Cr-large}     &     300M       &   300k        &     6.90      &    0.83  & 2.15 & 3.71 & 3.91 & 3.50 \\
\textsc{Zipa-Cr-small}     &     64M       &    500k       &    6.02       &   0.65  &   2.54 & 4.60 & 4.71 & 3.70    \\
\textsc{Zipa-Cr-large}     &     300M       &   500k        &      6.37     &   0.77  & 1.87 &  3.69  & 3.93 & 3.32 \\\midrule 
\textsc{Zipa-Cr-Ns-small}     &     64M       &    700k       &    \underline{5.94}       &   0.69  &   1.94 & 3.79 & \underline{3.87}  & \underline{3.24}   \\
\textsc{Zipa-Cr-Ns-large}     &     300M       &   800k        &  \textbf{5.93} & 0.75 &  \underline{1.75} & \underline{3.67} & 3.92  & \textbf{3.20} \\ \midrule
\textit{No diacritics$^{**}$} \\\midrule
\textsc{Zipa-Cr-Ns-small}     &     64M       &    700k       &   5.80       &   0.68  &   1.92 & 3.76 & 3.86  & 3.21   \\
\textsc{Zipa-Cr-Ns-large}     &     300M       &   780k        &  5.81 & 0.71 &  1.78 & 3.66 & 3.86  & 3.17 \\ 
\bottomrule       
\end{tabular}
}
\caption{Main PFER results on unseen languages and domains. $^{**}$Diacritics were removed for both training and evaluation sets, so results are not directly comparable with other models. \textbf{Notations:} \textsc{T} - Transducer; \textsc{Cr} - Consistency-regularized CTC; \textsc{Ns} - Noisy student training.}
\label{tab:unseen}
\end{table*}

\subsection{Baselines}
To contextualize the performance of the proposed model, we compared our models with several universal phone recognition models with publicly available weights. 
\begin{itemize}
    \setlength{\parskip}{1pt}  
    \item \textbf{Allosaurus}. Allosaurus \cite{li2020universal} is one of the earliest universal phone recognizers. The network backbone consists of bi-directional LSTM networks, and it has a shared phone output layers and language-specific allophone layers. 
    \item \textbf{Wav2Vec2Phoneme}. Wav2Vec2Phoneme \cite{xu22b_interspeech} is a state-of-the-art phone recognizer based on the pretrained Wav2Vec2 and XLSR-53 model \cite{baevski2020wav2vec,conneau2020unsupervised}. It was fine-tuned on 57k hours of speech data with phone transcriptions. We examined two checkpoints: \texttt{W2V2P-lv-60-ft}\footnote{\url{https://huggingface.co/facebook/wav2vec2-lv-60-espeak-cv-ft}}, which was fine-tuned from Wav2Vec2, and \texttt{W2V2P-xlsr-53-ft}\footnote{\url{https://huggingface.co/facebook/wav2vec2-xlsr-53-espeak-cv-ft}}, which was initialized with XLSR-53. 
    \item \textbf{MultIPA}. MultIPA\footnote{\url{https://huggingface.co/ctaguchi/wav2vec2-large-xlsr-japlmthufielta-ipa1000-ns}} \cite{taguchi23_interspeech} is another model based on the XLSR-53 \cite{conneau2020unsupervised}. The model was fine-tuned on relatively small but high-quality data with phone transcriptions, achieving competitive performance in phone recognition. 
    \item \textbf{Whisper-PPT}. Whisper-PPT \cite{samir2024efficiently} is an autoregressive universal phone recognition model based on the pretrained \texttt{Whisper-small} \cite{radford2023robust}. It was fine-tuned on a selected high-quality subset of IPAPack \cite{zhu-etal-2024-taste}. Unlike other models, the autoregressive nature of Whisper makes it uniquely prone to repeatedly generating hallucinated substrings on occasion.  
\end{itemize}
Allophant \cite{glocker23_interspeech} is another state-of-the-art phone recognizer based on XLS-R \cite{babu22_interspeech}. However, Allophant relies on an existing phoneset to make predictions. Some of our evaluation datasets, such as unseen languages and L2 speech, do not have an existing phoneset in PHOIBLE, so we did not compare with Allophant.

\section{Results}
We evaluated model performance with the PFER, which measures the alignment of binary articulatory features. The metric was computed with PanPhon \cite{mortensen-etal-2016-panphon}. The main results are presented in Table~\ref{tab:seen} and Table~\ref{tab:unseen}. Below we summarize our main findings. 

\paragraph{ZIPA models reach state-of-the-art performance on multilingual phone recognition.} We trained \textsc{Zipa} variants on 17k hours of multilingual data from scratch. Even the small \textsc{Zipa} models with only 64M parameters can outperform the 300M transformer baselines that have been pretrained and/or fine-tuned on much more data. For example, both \texttt{FAIR-lv-60-ft} and \texttt{FAIR-xlsr-60-ft} \cite{xu22b_interspeech} were initialized from pretrained weights and fine-tuned on 57k labeled data. Meanwhile, the Zipformer backbone is also much more memory efficient and less computationally intensive than the vanilla transformer in XLSR series of models and Whisper \cite{yao2023zipformer}. Our study shows that careful curation of data, including increasing data quantity and carefully normalizing the IPA labels, as well as a good choice of backbone model can yield effective improvement.  

\paragraph{Smaller models and non-autoregressive models generalize better to unseen languages but perform worse on seen languages.} 

Our results show that transducer models tend to outperform CTC based models on seen languages (see Table~\ref{tab:seen}). Autoregressive transducers model the dependencies better than CTC models, where conditional independence between labels is learned. However, learning the causal dependencies between phones can also hurt the multilingual generalizability, as unseen languages might have a different phonological structure. 
Larger models also tend to overfit the training data, weakening their abilities to predict unseen languages. This is particularly evident on both \texttt{Doreco} and the \texttt{VoxAngeles} test sets.  

Yet CTC models are still valuable as they are more efficient than autoregressive models and can be combined with an external alignment algorithm to generate approximate time stamps for multilingual data \cite{kurzinger2020ctc, pratap2024scaling}.

It is important to note that the evaluation metric PFER is a distance function rather than a ratio, so its magnitude tends to correlate with length. While it appears that the PFER for seen languages (Table~\ref{tab:seen}) is higher than the PFER for unseen languages (Table~\ref{tab:unseen}), it is because the speech samples from seen languages are longer than those from unseen languages. In Table~\ref{tab:seen}, some languages, especially \texttt{por} and \texttt{fre}, have consistently lower scores than other languages. This is caused by both the length of the evaluation samples and the phone set mismatch in these languages. For example, Portuguese uses [\textipa{5}] frequently but it is often transcribed as the more crosslinguistically frequent [a]. French marks nasality with a diacritic [\textasciitilde], but other languages tend to use the nasal consonants. Such mismatches in the phone set pose challenges to the phone recognition system, especially for small models. Yet longer training time and more model parameters enable models to memorize these language-specific conventions better. At least for high-resource languages, ZIPA models can implicitly distinguish the language and transcribe phones accordingly.

\begin{table}[]
\centering
\small
\begin{tabular}{lrr}
\toprule
Model & Seen Avg. & Unseen Avg. \\\midrule
\textsc{Zipa-Cr-l} \\
\hspace{0.2in} - 800k &   3.18        &    3.28         \\
\textsc{Zipa-Cr-Ns-l}     &           &    \\
\hspace{0.2in} - $\lambda$=0.5 & \textbf{2.71} & 3.20 \\
\hspace{0.2in} - $\lambda$=1.0 & 2.77 & \textbf{3.19} \\\midrule
\textsc{Zipa-Cr-Ns-s}     &           &            \\
\hspace{0.2in} - $\lambda$=0.2 &  3.36 & 3.35\\
\hspace{0.2in} - $\lambda$=0.5 & \textbf{3.22} & 3.24\\
\hspace{0.2in} - $\lambda$=1.0 & 3.41 & \textbf{3.23} \\\bottomrule
\end{tabular}
\caption{Ablation analysis of noisy student training.}
\label{tab: abation}
\end{table}

\paragraph{Noisy student training brings minor but consistent improvement.} \textsc{Zipa-Cr-Ns} models can consistently improve model performance, though the improvement is minor. This is likely because the pseudo-labels on unseen languages are extremely noisy, diminishing the benefits of additional data. Our ablation analysis in Table~\ref{tab: abation} indicates that continuing to train together with pseudo-labelled data is more beneficial than continuing to train on the existing labeled data beyond 500k steps. The value of $\lambda$ seems to control the trade-off between in-domain and out-of-domain performance, but the overall impact of $\lambda$ is not large. Note that we only adopted a simple approach to do noisy student training, so there is still room for improvement. Further research is needed to investigate how to better exploit the massive amount of unlabelled data.

\begin{figure*}
    \centering
    \includegraphics[width=\linewidth]{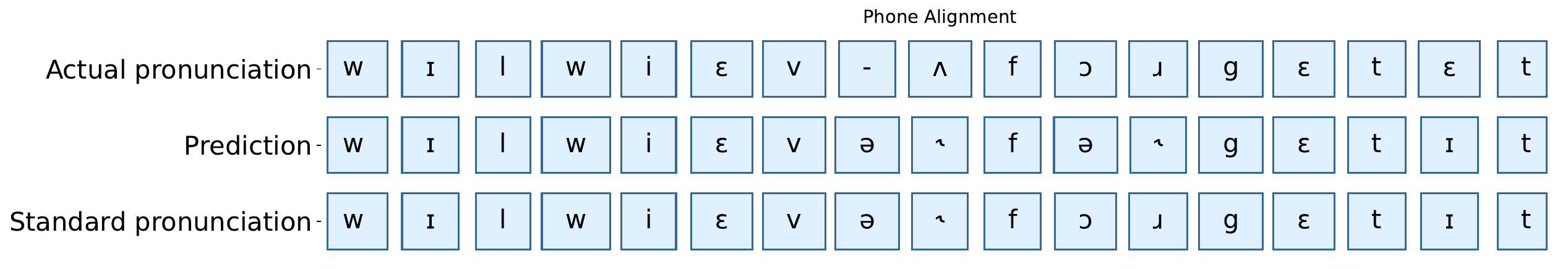}
    \caption{A sample transcription of the prompt ``\textit{Will we ever forget it}" in L2 speech by \textsc{Zipa-Cr-Ns-l}. The predicted transcription aligns more with the standard pronunciation, suggesting that the model failed to capture the actual sociophonetic variation. }
    \label{fig:alignment}
\end{figure*}

\begin{figure}
    \centering
    \includegraphics[width=\linewidth]{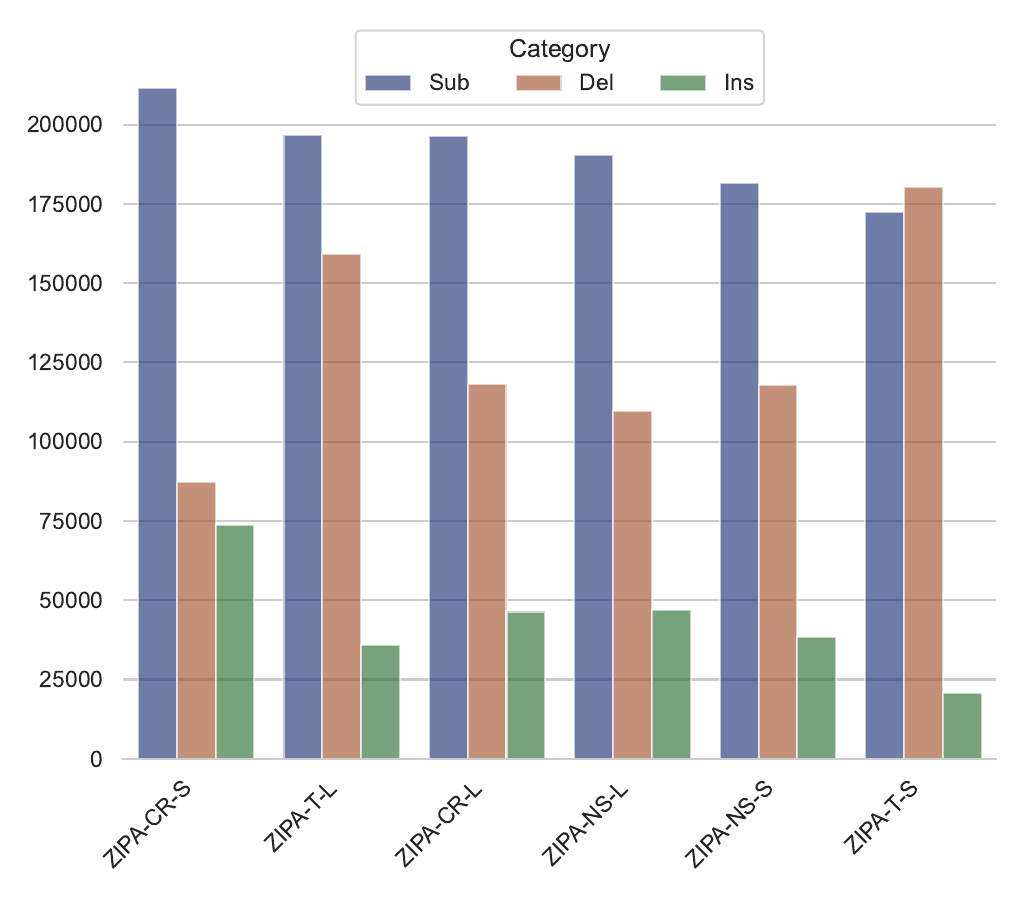}
    \caption{Distributions of transcription error types. Substitution errors are most common across models. Transducers exhibit a relatively high rate of deletion errors.}
    \label{fig:error_type}
\end{figure}

\begin{table}
\small
\centering
\begin{minipage}{0.3\linewidth}
    \begin{tabular}{cc}
    \toprule
    \textbf{IPA} & \textbf{Del} \\
    \midrule
    \textipa{:} & 17225 \\
    \textipa{P} & 11105 \\
    \textipa{i} & 7172 \\
    \textipa{a} & 5631 \\
    \textipa{n} & 4714 \\
    \textipa{h} & 4204 \\
    \textipa{e} & 3727 \\
    \textipa{E} & 3646 \\
    \textipa{u} & 3399 \\
     \textasciitilde & 3249 \\
    \textipa{o} & 3232 \\
    \textipa{t} & 2492 \\
    \textipa{@} & 2291 \\
    \textipa{I} & 2041 \\
    \textipa{w} & 1965 \\
    \textipa{j} & 1938 \\
    \textipa{d} & 1811 \\
    \textipa{k} & 1647 \\
    \textipa{O} & 1612 \\
    \textipa{\super h} & 1505 \\
    \bottomrule
    \end{tabular}
\end{minipage}%
\hspace{0.3cm} 
\begin{minipage}{0.3\linewidth}
    \begin{tabular}{cc}
    \toprule
    \textbf{IPA} & \textbf{Ins} \\
    \midrule
    \textipa{:} & 9777 \\
    \textipa{\|[c}$^*$ & 3924 \\
    \textipa{a} & 3401 \\
    \textipa{j} & 2793 \\
    \textipa{i} & 2491 \\
    \textipa{u} & 2123 \\
    \textipa{e} & 1943 \\
    \textipa{t} & 1487 \\
    \textipa{k} & 1487 \\
    \textipa{S} & 1414 \\
    \textipa{Z} & 1289 \\
    \textipa{n} & 1265 \\
    \textipa{U} & 894 \\
    \textipa{o} & 814 \\
    \textipa{p} & 781 \\
    \textipa{w} & 759 \\
    \textipa{@} & 696 \\
    \textipa{r} & 667 \\
    \textipa{d} & 653 \\
    \textipa{R} & 648 \\
    \bottomrule
    \end{tabular}
\end{minipage}%
\hspace{0.3cm} 
\begin{minipage}{0.3\linewidth}
    \begin{tabular}{cc}
    \toprule
    \textbf{IPA} & \textbf{Sub} \\
    \midrule
    \textipa{a} $\to$ \textipa{A} & 5669 \\
    \textipa{i} $\to$ \textipa{e} & 5592 \\
    \textipa{E} $\to$ \textipa{e} & 4454 \\
    \textipa{o} $\to$ \textipa{u} & 3478 \\
    \textipa{e} $\to$ \textipa{i} & 3089 \\
    \textipa{u} $\to$ \textipa{o} & 2889 \\
    \textipa{O} $\to$ \textipa{o} & 2678 \\
    \textipa{E} $\to$ \textipa{a} & 2480 \\
    \textipa{@} $\to$ \textipa{a} & 2226 \\
    \textipa{e} $\to$ \textipa{a} & 1932 \\
    \textipa{b} $\to$ \textipa{p} & 1859 \\
    \textipa{d} $\to$ \textipa{t} & 1717 \\
    \textipa{o} $\to$ \textipa{O} & 1716 \\
    \textipa{i} $\to$ \textipa{j} & 1619 \\
    \textipa{g} $\to$ \textipa{k} & 1609 \\
    \textipa{o} $\to$ \textipa{a} & 1526 \\
    \textipa{i} $\to$ \textipa{I} & 1444 \\
    \textipa{E} $\to$ \textipa{e} & 1436 \\
    \textipa{r} $\to$ \textipa{R} & 1429 \\
    \textipa{e} $\to$ \textipa{E} & 1425 \\
    \bottomrule
    \end{tabular}
\end{minipage}
\caption{Summary of \textbf{Del}etions, \textbf{Ins}ertions, and \textbf{Sub}stitution errors by \textsc{Zipa-Cr-Ns-L}. Other \textsc{Zipa} models also exhibit a similar pattern. $^*$c denotes any consonant. }
\label{tab:errors}
\end{table}

\paragraph{Removing diacritics can improve the match between model predictions and ground truth, especially on unseen languages, but the impact is slight.} Both Table~\ref{tab:seen} and~\ref{tab:unseen} suggest that the no-diacritic condition yields inconsistent and slight improvement, as the number of total symbols is reduced. Our further inspection indicates that ZIPA models tend to handle diacritics pretty well for seen languages, as the patterns in these languages are probably well memorized during training. Yet, generalizing diacritics across languages poses a much larger challenge. The largest change in score is the \texttt{Doreco} evaluation set, as it contains more diacritics than other datasets \citep{paschen2020building}.

\section{Analysis} \label{sec:analysis}
We also conducted an error analysis to understand model behaviors and present findings below.
\paragraph{Phone recognition models tend to smooth out the phonetic variation during inference.}  In Table~\ref{tab:unseen}, there is a systematic gap between the performance of \texttt{L2-Standard} and the \texttt{L2-Perceived} test sets. In Figure~\ref{fig:alignment}, given the exact same L2 speech,  \textsc{Zipa} predictions tend to better match the standard dictionary pronunciation than the actual pronunciation. This is likely an artifact of data curation, as all of the training data were generated from pronunciation dictionaries and G2P models. Yet this finding also implies that the phone recognition models are still matching the frequent phone patterns in the dataset, rather than transcribing phones as they are actually produced. 

\paragraph{Vowels are more difficult to predict crosslingusitically.} Prior research has revealed that certain sounds are recognized better across languages \cite{zelasko2022discovering}. We conducted an error analysis of the model predictions on \texttt{Doreco}. As shown in Figure~\ref{fig:error_type}, substitution errors are far more common than addition and deletion errors. Transducer models show much higher deletion errors than CTC models. Our close inspection also suggests that transducers generate quite a few empty transcriptions for unseen languages.  

Table~\ref{tab:errors} provides further details on the top errors made by \textsc{Zipa-Cr-Ns}. The top deletion and insertion errors are diacritics. The length symbol \textipa{:} is consistently the most frequently added or deleted symbol, as vowel length is relative across languages. The glottal stop \textipa{P} is often not contrastive and not explicitly marked in IPA transcriptions, resulting in high deletions in model predictions. For substitution, the top errors are the substitution of vowels that are close in the vowel space. Compared to consonants, vowel realizations tend to be more gradient in their acoustics, resulting in higher acoustic overlap between otherwise contrastive vowel categories and therefore more ambiguous. Such misidentification patterns also mirror the patterns of human speech perception crosslinguistically \cite{sebastian2005cross}. 




\section{Conclusions}
In conclusion, we present a large-scale multilingual phone recognition dataset \textsc{Ipa Pack++} and a series of Zipformer-based \textsc{Zipa} models, which exhibit state-of-the-art performance on phone recognition. We hope that our research can provide foundations to support more downstream multilingual speech processing tasks that benefit from phonetic transcriptions. Yet simply scaling up the G2P for transcribed speech data alone might not be able to solve phone recognition, as models can simply memorize the standard pronunciation. We will also actively explore how to incorporate more linguistic knowledge to further improve performance.

\section*{Ethics statement}
We adhere to ethical practices in our research. We only selected publicly available datasets with permissive licenses that allow us to redistribute the processed data and the models. We believe that open-sourcing our research can help facilitate future research towards multilingual speech technologies for both the speech processing communities and the linguistics communities.

It is our firm belief that this research can contribute to the promotion of more inclusive speech technologies for more languages, especially for under-represented languages. While our model is primarily developed to support language documentation and other downstream applications, we are also aware that multilingual speech recognition can exhibit biases towards non-mainstream accents and potentially be used for malicious purposes such as surveillance. We urge that caution be exercised when deploying such models in downstream tasks. 

\section*{Limitations}
Our study is still limited in several ways. First, the number of languages studied in our paper is still limited. The distribution of languages is highly skewed in our dataset, which still biases our models towards high-resource languages. 

Secondly, our current approach trains models on synthetic labels from G2P. However, the data quality is limited as dictionary pronunciations might not reflect the actual pronunciation in spontaneous speech. This also results in the \textsc{Zipa} models to smooth out variation in some L2 speech. More research is needed to investigate how to curate higher quality data for phone recognition that can reflect the actual pronunciation. 

The limitation of computational resources also limits our abilities to perform extensive hyperparameter tuning and conduct extensive experiments to explore different architectures and pseudo-labeling strategies. In the future, we will continue to explore better strategies to continue to improve the performance of multilingual speech processing systems. 

\section*{Acknowledgments}
This research was enabled in part through the computational resources provided by Advanced Research Computing at the University of British Columbia and the Digital Research Alliance of Canada. FS is supported by an NSERC PGS-D Scholarship. The research activities were also supported by the NSERC Discovery Grant and the CFI JELF Grant awarded to JZ and by SNF Grant  PR00P1\_208460 to EC.

\bibliography{custom,anthology}

\begin{thebibliography}{73}
\providecommand{\natexlab}[1]{#1}

\bibitem[{Abramson and Whalen(2017)}]{abramson2017voice}
Arthur~S Abramson and Douglas~H Whalen. 2017.
\newblock Voice onset time (vot) at 50: Theoretical and practical issues in measuring voicing distinctions.
\newblock \emph{Journal of phonetics}, 63:75--86.

\bibitem[{Anderson et~al.(2023)Anderson, Tresoldi, Greenhill, Forkel, Gray, and List}]{anderson2023variation}
Cormac Anderson, Tiago Tresoldi, Simon~J Greenhill, Robert Forkel, Russell Gray, and Johann-Mattis List. 2023.
\newblock Variation in phoneme inventories: quantifying the problem and improving comparability.
\newblock \emph{Journal of Language Evolution}, page lzad011.

\bibitem[{Ardila et~al.(2020)Ardila, Branson, Davis, Kohler, Meyer, Henretty, Morais, Saunders, Tyers, and Weber}]{ardila-etal-2020-common}
Rosana Ardila, Megan Branson, Kelly Davis, Michael Kohler, Josh Meyer, Michael Henretty, Reuben Morais, Lindsay Saunders, Francis Tyers, and Gregor Weber. 2020.
\newblock \href {https://aclanthology.org/2020.lrec-1.520/} {Common voice: A massively-multilingual speech corpus}.
\newblock In \emph{Proceedings of the Twelfth Language Resources and Evaluation Conference}, pages 4218--4222, Marseille, France. European Language Resources Association.

\bibitem[{Babu et~al.(2022)Babu, Wang, Tjandra, Lakhotia, Xu, Goyal, Singh, {von Platen}, Saraf, Pino, Baevski, Conneau, and Auli}]{babu22_interspeech}
Arun Babu, Changhan Wang, Andros Tjandra, Kushal Lakhotia, Qiantong Xu, Naman Goyal, Kritika Singh, Patrick {von Platen}, Yatharth Saraf, Juan Pino, Alexei Baevski, Alexis Conneau, and Michael Auli. 2022.
\newblock \href {https://doi.org/10.21437/Interspeech.2022-143} {Xls-r: Self-supervised cross-lingual speech representation learning at scale}.
\newblock In \emph{Interspeech 2022}, pages 2278--2282.

\bibitem[{Babu et~al.(2021)Babu, Wang, Tjandra, Lakhotia, Xu, Goyal, Singh, von Platen, Saraf, Pino et~al.}]{babu2021xls}
Arun Babu, Changhan Wang, Andros Tjandra, Kushal Lakhotia, Qiantong Xu, Naman Goyal, Kritika Singh, Patrick von Platen, Yatharth Saraf, Juan Pino, et~al. 2021.
\newblock Xls-r: Self-supervised cross-lingual speech representation learning at scale.
\newblock \emph{arXiv preprint arXiv:2111.09296}.

\bibitem[{Baevski et~al.(2020)Baevski, Zhou, Mohamed, and Auli}]{baevski2020wav2vec}
Alexei Baevski, Yuhao Zhou, Abdelrahman Mohamed, and Michael Auli. 2020.
\newblock wav2vec 2.0: A framework for self-supervised learning of speech representations.
\newblock \emph{Advances in neural information processing systems}, 33:12449--12460.

\bibitem[{Bu et~al.(2017)Bu, Du, Na, Wu, and Zheng}]{bu2017aishell}
Hui Bu, Jiayu Du, Xingyu Na, Bengu Wu, and Hao Zheng. 2017.
\newblock Aishell-1: An open-source mandarin speech corpus and a speech recognition baseline.
\newblock In \emph{2017 20th conference of the oriental chapter of the international coordinating committee on speech databases and speech I/O systems and assessment (O-COCOSDA)}, pages 1--5. IEEE.

\bibitem[{Chen et~al.(2024)Chen, Zhang, Peng, Li, Tian, Shi, Chang, Maiti, Livescu, and Watanabe}]{chen-etal-2024-towards-robust}
William Chen, Wangyou Zhang, Yifan Peng, Xinjian Li, Jinchuan Tian, Jiatong Shi, Xuankai Chang, Soumi Maiti, Karen Livescu, and Shinji Watanabe. 2024.
\newblock \href {https://doi.org/10.18653/v1/2024.emnlp-main.570} {Towards robust speech representation learning for thousands of languages}.
\newblock In \emph{Proceedings of the 2024 Conference on Empirical Methods in Natural Language Processing}, pages 10205--10224, Miami, Florida, USA. Association for Computational Linguistics.

\bibitem[{Cho and Ladefoged(1999)}]{cho1999variation}
Taehong Cho and Peter Ladefoged. 1999.
\newblock Variation and universals in vot: evidence from 18 languages.
\newblock \emph{Journal of phonetics}, 27(2):207--229.

\bibitem[{Chodroff et~al.(2019)Chodroff, Golden, and Wilson}]{chodroff2019covariation}
Eleanor Chodroff, Alessandra Golden, and Colin Wilson. 2019.
\newblock Covariation of stop voice onset time across languages: Evidence for a universal constraint on phonetic realization.
\newblock \emph{The Journal of the Acoustical Society of America}, 145(1):EL109--EL115.

\bibitem[{Chodroff et~al.(2024)Chodroff, Pa{\v{z}}on, Baker, and Moran}]{chodroff-etal-2024-phonetic}
Eleanor Chodroff, Bla{\v{z}} Pa{\v{z}}on, Annie Baker, and Steven Moran. 2024.
\newblock \href {https://aclanthology.org/2024.lrec-main.1114/} {Phonetic segmentation of the {UCLA} phonetics lab archive}.
\newblock In \emph{Proceedings of the 2024 Joint International Conference on Computational Linguistics, Language Resources and Evaluation (LREC-COLING 2024)}, pages 12724--12733, Torino, Italia. ELRA and ICCL.

\bibitem[{Conneau et~al.(2020)Conneau, Baevski, Collobert, Mohamed, and Auli}]{conneau2020unsupervised}
Alexis Conneau, Alexei Baevski, Ronan Collobert, Abdelrahman Mohamed, and Michael Auli. 2020.
\newblock Unsupervised cross-lingual representation learning for speech recognition.
\newblock \emph{arXiv preprint arXiv:2006.13979}.

\bibitem[{Feng et~al.(2023)Feng, Tu, Xia, Huang, and Wang}]{feng23b_interspeech}
Siyuan Feng, Ming Tu, Rui Xia, Chuanzeng Huang, and Yuxuan Wang. 2023.
\newblock \href {https://doi.org/10.21437/Interspeech.2023-621} {Language-universal phonetic encoder for low-resource speech recognition}.
\newblock In \emph{Interspeech 2023}, pages 1429--1433.

\bibitem[{Fuchs et~al.(2022)Fuchs, Hoshen, and Keshet}]{fuchs22_interspeech}
Tzeviya Fuchs, Yedid Hoshen, and Yossi Keshet. 2022.
\newblock \href {https://doi.org/10.21437/Interspeech.2022-11474} {Unsupervised word segmentation using k nearest neighbors}.
\newblock In \emph{Interspeech 2022}, pages 4646--4650.

\bibitem[{Gao et~al.(2021)Gao, Ni, Zhang, Qian, Chang, and Hasegawa-Johnson}]{gao2021zero}
Heting Gao, Junrui Ni, Yang Zhang, Kaizhi Qian, Shiyu Chang, and Mark Hasegawa-Johnson. 2021.
\newblock Zero-shot cross-lingual phonetic recognition with external language embedding.
\newblock In \emph{Interspeech}, pages 1304--1308.

\bibitem[{Ghodsi et~al.(2020)Ghodsi, Liu, Apfel, Cabrera, and Weinstein}]{ghodsi2020rnn}
Mohammadreza Ghodsi, Xiaofeng Liu, James Apfel, Rodrigo Cabrera, and Eugene Weinstein. 2020.
\newblock Rnn-transducer with stateless prediction network.
\newblock In \emph{ICASSP 2020-2020 IEEE International Conference on Acoustics, Speech and Signal Processing (ICASSP)}, pages 7049--7053. IEEE.

\bibitem[{Glocker et~al.(2023)Glocker, Herygers, and Georges}]{glocker23_interspeech}
Kevin Glocker, Aaricia Herygers, and Munir Georges. 2023.
\newblock \href {https://doi.org/10.21437/Interspeech.2023-772} {Allophant: Cross-lingual phoneme recognition with articulatory attributes}.
\newblock In \emph{INTERSPEECH 2023}, pages 2258--2262.

\bibitem[{Gong et~al.(2022)Gong, Chen, Chu, Chang, and Glass}]{gong2022transformer}
Yuan Gong, Ziyi Chen, Iek-Heng Chu, Peng Chang, and James Glass. 2022.
\newblock Transformer-based multi-aspect multi-granularity non-native english speaker pronunciation assessment.
\newblock In \emph{ICASSP 2022-2022 IEEE International Conference on Acoustics, Speech and Signal Processing (ICASSP)}, pages 7262--7266. IEEE.

\bibitem[{Graves et~al.(2006)Graves, Fern{\'a}ndez, Gomez, and Schmidhuber}]{graves2006connectionist}
Alex Graves, Santiago Fern{\'a}ndez, Faustino Gomez, and J{\"u}rgen Schmidhuber. 2006.
\newblock Connectionist temporal classification: labelling unsegmented sequence data with recurrent neural networks.
\newblock In \emph{Proceedings of the 23rd international conference on Machine learning}, pages 369--376.

\bibitem[{Gulati et~al.(2020)Gulati, Qin, Chiu, Parmar, Zhang, Yu, Han, Wang, Zhang, Wu, and Pang}]{gulati20_interspeech}
Anmol Gulati, James Qin, Chung-Cheng Chiu, Niki Parmar, Yu~Zhang, Jiahui Yu, Wei Han, Shibo Wang, Zhengdong Zhang, Yonghui Wu, and Ruoming Pang. 2020.
\newblock \href {https://doi.org/10.21437/Interspeech.2020-3015} {Conformer: Convolution-augmented transformer for speech recognition}.
\newblock In \emph{Interspeech 2020}, pages 5036--5040.

\bibitem[{Hwang et~al.(2022{\natexlab{a}})Hwang, Misra, Huo, Siddhartha, Garg, Qiu, Sim, Strohman, Beaufays, and He}]{hwang2022large}
Dongseong Hwang, Ananya Misra, Zhouyuan Huo, Nikhil Siddhartha, Shefali Garg, David Qiu, Khe~Chai Sim, Trevor Strohman, Fran{\c{c}}oise Beaufays, and Yanzhang He. 2022{\natexlab{a}}.
\newblock Large-scale asr domain adaptation using self-and semi-supervised learning.
\newblock In \emph{ICASSP 2022-2022 IEEE International Conference on Acoustics, Speech and Signal Processing (ICASSP)}, pages 6627--6631. IEEE.

\bibitem[{Hwang et~al.(2022{\natexlab{b}})Hwang, Sim, Huo, and Strohman}]{hwang22c_interspeech}
Dongseong Hwang, Khe~Chai Sim, Zhouyuan Huo, and Trevor Strohman. 2022{\natexlab{b}}.
\newblock \href {https://doi.org/10.21437/Interspeech.2022-11034} {Pseudo label is better than human label}.
\newblock In \emph{Interspeech 2022}, pages 1421--1425.

\bibitem[{International Phonetic~Association(1999)}]{international1999handbook}
IPA International Phonetic~Association. 1999.
\newblock \emph{Handbook of the International Phonetic Association: A guide to the use of the International Phonetic Alphabet}.
\newblock Cambridge University Press.

\bibitem[{Kerswill and Wright(1990)}]{kerswill1990validity}
Paul Kerswill and Susan Wright. 1990.
\newblock The validity of phonetic transcription: Limitations of a sociolinguistic research tool.
\newblock \emph{Language variation and change}, 2(3):255--275.

\bibitem[{Khassanov et~al.(2021)Khassanov, Mussakhojayeva, Mirzakhmetov, Adiyev, Nurpeiissov, and Varol}]{khassanov-etal-2021-crowdsourced}
Yerbolat Khassanov, Saida Mussakhojayeva, Almas Mirzakhmetov, Alen Adiyev, Mukhamet Nurpeiissov, and Huseyin~Atakan Varol. 2021.
\newblock \href {https://doi.org/10.18653/v1/2021.eacl-main.58} {A crowdsourced open-source {K}azakh speech corpus and initial speech recognition baseline}.
\newblock In \emph{Proceedings of the 16th Conference of the European Chapter of the Association for Computational Linguistics: Main Volume}, pages 697--706, Online. Association for Computational Linguistics.

\bibitem[{Kim et~al.(2023)Kim, Wu, Peng, Pan, Sridhar, Han, and Watanabe}]{kim2023branchformer}
Kwangyoun Kim, Felix Wu, Yifan Peng, Jing Pan, Prashant Sridhar, Kyu~J Han, and Shinji Watanabe. 2023.
\newblock E-branchformer: Branchformer with enhanced merging for speech recognition.
\newblock In \emph{2022 IEEE Spoken Language Technology Workshop (SLT)}, pages 84--91. IEEE.

\bibitem[{Kjartansson et~al.(2018)Kjartansson, Sarin, Pipatsrisawat, Jansche, and Ha}]{kjartansson-etal-sltu2018}
Oddur Kjartansson, Supheakmungkol Sarin, Knot Pipatsrisawat, Martin Jansche, and Linne Ha. 2018.
\newblock \href {http://dx.doi.org/10.21437/SLTU.2018-11} {{Crowd-Sourced Speech Corpora for Javanese, Sundanese, Sinhala, Nepali, and Bangladeshi Bengali}}.
\newblock In \emph{Proc. The 6th Intl. Workshop on Spoken Language Technologies for Under-Resourced Languages (SLTU)}, pages 52--55, Gurugram, India.

\bibitem[{Kuang et~al.(2022)Kuang, Guo, Kang, Lin, Luo, Yao, and Povey}]{kuang22_interspeech}
Fangjun Kuang, Liyong Guo, Wei Kang, Long Lin, Mingshuang Luo, Zengwei Yao, and Daniel Povey. 2022.
\newblock \href {https://doi.org/10.21437/Interspeech.2022-10340} {Pruned rnn-t for fast, memory-eﬀicient asr training}.
\newblock In \emph{Interspeech 2022}, pages 2068--2072.

\bibitem[{K{\"u}rzinger et~al.(2020)K{\"u}rzinger, Winkelbauer, Li, Watzel, and Rigoll}]{kurzinger2020ctc}
Ludwig K{\"u}rzinger, Dominik Winkelbauer, Lujun Li, Tobias Watzel, and Gerhard Rigoll. 2020.
\newblock Ctc-segmentation of large corpora for german end-to-end speech recognition.
\newblock In \emph{International Conference on Speech and Computer}, pages 267--278. Springer.

\bibitem[{Ladefoged and Johnson(2014)}]{ladefoged2014course}
Peter Ladefoged and Keith Johnson. 2014.
\newblock \emph{A course in phonetics}.
\newblock Cengage learning.

\bibitem[{Lane and Bird(2021)}]{lane2021local}
William Lane and Steven Bird. 2021.
\newblock Local word discovery for interactive transcription.
\newblock In \emph{Proceedings of the 2021 Conference on Empirical Methods in Natural Language Processing}, pages 2058--2067.

\bibitem[{Lee et~al.(2022)Lee, Noh, Nam, and Lee}]{lee2022duration}
Sang-Hoon Lee, Hyeong-Rae Noh, Woo-Jeoung Nam, and Seong-Whan Lee. 2022.
\newblock Duration controllable voice conversion via phoneme-based information bottleneck.
\newblock \emph{IEEE/ACM Transactions on Audio, Speech, and Language Processing}, 30:1173--1183.

\bibitem[{Li(2017)}]{li2017divination}
Xiaochang Li. 2017.
\newblock \emph{Divination engines: A media history of text prediction}.
\newblock Ph.D. thesis, New York University.

\bibitem[{Li et~al.(2020)Li, Dalmia, Li, Lee, Littell, Yao, Anastasopoulos, Mortensen, Neubig, Black et~al.}]{li2020universal}
Xinjian Li, Siddharth Dalmia, Juncheng Li, Matthew Lee, Patrick Littell, Jiali Yao, Antonios Anastasopoulos, David~R Mortensen, Graham Neubig, Alan~W Black, et~al. 2020.
\newblock Universal phone recognition with a multilingual allophone system.
\newblock In \emph{ICASSP 2020-2020 IEEE International Conference on Acoustics, Speech and Signal Processing (ICASSP)}, pages 8249--8253. IEEE.

\bibitem[{Li et~al.(2021)Li, Mortensen, Metze, and Black}]{li2021multilingual}
Xinjian Li, David~R Mortensen, Florian Metze, and Alan~W Black. 2021.
\newblock Multilingual phonetic dataset for low resource speech recognition.
\newblock In \emph{ICASSP 2021-2021 IEEE International Conference on Acoustics, Speech and Signal Processing (ICASSP)}, pages 6958--6962. IEEE.

\bibitem[{Liu et~al.(2023)Liu, Ling, and Chen}]{liu2023pronunciation}
Chang Liu, Zhen-Hua Ling, and Ling-Hui Chen. 2023.
\newblock Pronunciation dictionary-free multilingual speech synthesis using learned phonetic representations.
\newblock \emph{IEEE/ACM Transactions on Audio, Speech, and Language Processing}.

\bibitem[{Liu et~al.(2021)Liu, Peng, Xiong, and Lu}]{liu2021phoneme}
Yajing Liu, Xiulian Peng, Zhiwei Xiong, and Yan Lu. 2021.
\newblock Phoneme-based distribution regularization for speech enhancement.
\newblock In \emph{ICASSP 2021-2021 IEEE International Conference on Acoustics, Speech and Signal Processing (ICASSP)}, pages 726--730. IEEE.

\bibitem[{Madhavaraj et~al.(2022{\natexlab{a}})Madhavaraj, Pilar, and G}]{mile_2}
A~Madhavaraj, Bharathi Pilar, and Ramakrishnan~A G. 2022{\natexlab{a}}.
\newblock \href {https://doi.org/10.48550/ARXIV.2207.13333} {Knowledge-driven subword grammar modeling for automatic speech recognition in tamil and kannada}.
\newblock \emph{arXiv preprint}.

\bibitem[{Madhavaraj et~al.(2022{\natexlab{b}})Madhavaraj, Pilar, and G}]{mile_1}
A~Madhavaraj, Bharathi Pilar, and Ramakrishnan~A G. 2022{\natexlab{b}}.
\newblock \href {https://doi.org/10.48550/ARXIV.2207.13331} {Subword dictionary learning and segmentation techniques for automatic speech recognition in tamil and kannada}.
\newblock \emph{arXiv preprint}.

\bibitem[{Mansurova and Kadyrbek(2023)}]{mansurova-kadyrbek-2023-kazakh-speech-dataset}
Madina Mansurova and Nurgali Kadyrbek. 2023.
\newblock \href {https://doi.org/10.3390/bdcc7030132} {The development of a kazakh speech recognition model using a convolutional neural network with fixed character level filters}.
\newblock In \emph{Proceedings of the Big Data and Cognitive Computing}, pages 5--9.

\bibitem[{Moran et~al.(2014)Moran, McCloy, and Wright}]{moran2014phoible}
Steven Moran, Daniel McCloy, and Richard Wright. 2014.
\newblock Phoible online.

\bibitem[{Mortensen et~al.(2018)Mortensen, Dalmia, and Littell}]{mortensen-etal-2018-epitran}
David~R. Mortensen, Siddharth Dalmia, and Patrick Littell. 2018.
\newblock \href {https://aclanthology.org/L18-1429/} {{E}pitran: Precision {G}2{P} for many languages}.
\newblock In \emph{Proceedings of the Eleventh International Conference on Language Resources and Evaluation ({LREC} 2018)}, Miyazaki, Japan. European Language Resources Association (ELRA).

\bibitem[{Mortensen et~al.(2016)Mortensen, Littell, Bharadwaj, Goyal, Dyer, and Levin}]{mortensen-etal-2016-panphon}
David~R. Mortensen, Patrick Littell, Akash Bharadwaj, Kartik Goyal, Chris Dyer, and Lori Levin. 2016.
\newblock \href {https://aclanthology.org/C16-1328/} {{P}an{P}hon: A resource for mapping {IPA} segments to articulatory feature vectors}.
\newblock In \emph{Proceedings of {COLING} 2016, the 26th International Conference on Computational Linguistics: Technical Papers}, pages 3475--3484, Osaka, Japan. The COLING 2016 Organizing Committee.

\bibitem[{Panayotov et~al.(2015)Panayotov, Chen, Povey, and Khudanpur}]{panayotov2015librispeech}
Vassil Panayotov, Guoguo Chen, Daniel Povey, and Sanjeev Khudanpur. 2015.
\newblock Librispeech: an asr corpus based on public domain audio books.
\newblock In \emph{2015 IEEE international conference on acoustics, speech and signal processing (ICASSP)}, pages 5206--5210. IEEE.

\bibitem[{Park et~al.(2019)Park, Chan, Zhang, Chiu, Zoph, Cubuk, and Le}]{park19e_interspeech}
Daniel~S. Park, William Chan, Yu~Zhang, Chung-Cheng Chiu, Barret Zoph, Ekin~D. Cubuk, and Quoc~V. Le. 2019.
\newblock \href {https://doi.org/10.21437/Interspeech.2019-2680} {Specaugment: A simple data augmentation method for automatic speech recognition}.
\newblock In \emph{Interspeech 2019}, pages 2613--2617.

\bibitem[{Park et~al.(2020)Park, Zhang, Jia, Han, Chiu, Li, Wu, and Le}]{park20d_interspeech}
Daniel~S. Park, Yu~Zhang, Ye~Jia, Wei Han, Chung-Cheng Chiu, Bo~Li, Yonghui Wu, and Quoc~V. Le. 2020.
\newblock \href {https://doi.org/10.21437/Interspeech.2020-1470} {Improved noisy student training for automatic speech recognition}.
\newblock In \emph{Interspeech 2020}, pages 2817--2821.

\bibitem[{Paschen et~al.(2020)Paschen, Delafontaine, Draxler, Fuchs, Stave, and Seifart}]{paschen2020building}
Ludger Paschen, Fran{\c{c}}ois Delafontaine, Christoph Draxler, Susanne Fuchs, Matthew Stave, and Frank Seifart. 2020.
\newblock Building a time-aligned cross-linguistic reference corpus from language documentation data (doreco).
\newblock In \emph{Proceedings of the Twelfth Language Resources and Evaluation Conference}, pages 2657--2666.

\bibitem[{Peng et~al.(2022)Peng, Dalmia, Lane, and Watanabe}]{peng2022branchformer}
Yifan Peng, Siddharth Dalmia, Ian Lane, and Shinji Watanabe. 2022.
\newblock Branchformer: Parallel mlp-attention architectures to capture local and global context for speech recognition and understanding.
\newblock In \emph{International Conference on Machine Learning}, pages 17627--17643. PMLR.

\bibitem[{Peng et~al.(2024{\natexlab{a}})Peng, Sudo, Shakeel, and Watanabe}]{peng2024owsm}
Yifan Peng, Yui Sudo, Muhammad Shakeel, and Shinji Watanabe. 2024{\natexlab{a}}.
\newblock Owsm-ctc: An open encoder-only speech foundation model for speech recognition, translation, and language identification.
\newblock \emph{arXiv preprint arXiv:2402.12654}.

\bibitem[{Peng et~al.(2024{\natexlab{b}})Peng, Sudo, Shakeel, and Watanabe}]{peng-etal-2024-owsm}
Yifan Peng, Yui Sudo, Muhammad Shakeel, and Shinji Watanabe. 2024{\natexlab{b}}.
\newblock \href {https://doi.org/10.18653/v1/2024.acl-long.549} {{OWSM}-{CTC}: An open encoder-only speech foundation model for speech recognition, translation, and language identification}.
\newblock In \emph{Proceedings of the 62nd Annual Meeting of the Association for Computational Linguistics (Volume 1: Long Papers)}, pages 10192--10209, Bangkok, Thailand. Association for Computational Linguistics.

\bibitem[{Pirklbauer et~al.(2023)Pirklbauer, Sach, Fluyt, Tirry, Wardah, Moeller, and Fingscheidt}]{pirklbauer2023evaluation}
Jan Pirklbauer, Marvin Sach, Kristoff Fluyt, Wouter Tirry, Wafaa Wardah, Sebastian Moeller, and Tim Fingscheidt. 2023.
\newblock Evaluation metrics for generative speech enhancement methods: Issues and perspectives.
\newblock In \emph{Speech Communication; 15th ITG Conference}, pages 265--269. VDE.

\bibitem[{Pitt et~al.(2005)Pitt, Johnson, Hume, Kiesling, and Raymond}]{pitt2005buckeye}
Mark~A Pitt, Keith Johnson, Elizabeth Hume, Scott Kiesling, and William Raymond. 2005.
\newblock The buckeye corpus of conversational speech: Labeling conventions and a test of transcriber reliability.
\newblock \emph{Speech Communication}, 45(1):89--95.

\bibitem[{Pratap et~al.(2024)Pratap, Tjandra, Shi, Tomasello, Babu, Kundu, Elkahky, Ni, Vyas, Fazel-Zarandi et~al.}]{pratap2024scaling}
Vineel Pratap, Andros Tjandra, Bowen Shi, Paden Tomasello, Arun Babu, Sayani Kundu, Ali Elkahky, Zhaoheng Ni, Apoorv Vyas, Maryam Fazel-Zarandi, et~al. 2024.
\newblock Scaling speech technology to 1,000+ languages.
\newblock \emph{Journal of Machine Learning Research}, 25(97):1--52.

\bibitem[{Pratap et~al.(2020)Pratap, Xu, Sriram, Synnaeve, and Collobert}]{pratap20_interspeech}
Vineel Pratap, Qiantong Xu, Anuroop Sriram, Gabriel Synnaeve, and Ronan Collobert. 2020.
\newblock \href {https://doi.org/10.21437/Interspeech.2020-2826} {Mls: A large-scale multilingual dataset for speech research}.
\newblock In \emph{Interspeech 2020}, pages 2757--2761.

\bibitem[{Radford et~al.(2023)Radford, Kim, Xu, Brockman, McLeavey, and Sutskever}]{radford2023robust}
Alec Radford, Jong~Wook Kim, Tao Xu, Greg Brockman, Christine McLeavey, and Ilya Sutskever. 2023.
\newblock Robust speech recognition via large-scale weak supervision.
\newblock In \emph{International Conference on Machine Learning}, pages 28492--28518. PMLR.

\bibitem[{Ramirez et~al.(2024)Ramirez, Chkhetiani, Ehrenberg, McHardy, Botros, Khare, Vanzo, Peyash, Oexle, Liang et~al.}]{ramirez2024anatomy}
Francis~McCann Ramirez, Luka Chkhetiani, Andrew Ehrenberg, Robert McHardy, Rami Botros, Yash Khare, Andrea Vanzo, Taufiquzzaman Peyash, Gabriel Oexle, Michael Liang, et~al. 2024.
\newblock Anatomy of industrial scale multilingual asr.
\newblock \emph{arXiv preprint arXiv:2404.09841}.

\bibitem[{Ronneberger et~al.(2015)Ronneberger, Fischer, and Brox}]{ronneberger2015u}
Olaf Ronneberger, Philipp Fischer, and Thomas Brox. 2015.
\newblock U-net: Convolutional networks for biomedical image segmentation.
\newblock In \emph{Medical image computing and computer-assisted intervention--MICCAI 2015: 18th international conference, Munich, Germany, October 5-9, 2015, proceedings, part III 18}, pages 234--241. Springer.

\bibitem[{Salesky et~al.(2020)Salesky, Chodroff, Pimentel, Wiesner, Cotterell, Black, and Eisner}]{salesky-etal-2020-corpus}
Elizabeth Salesky, Eleanor Chodroff, Tiago Pimentel, Matthew Wiesner, Ryan Cotterell, Alan~W Black, and Jason Eisner. 2020.
\newblock \href {https://doi.org/10.18653/v1/2020.acl-main.415} {A corpus for large-scale phonetic typology}.
\newblock In \emph{Proceedings of the 58th Annual Meeting of the Association for Computational Linguistics}, pages 4526--4546, Online. Association for Computational Linguistics.

\bibitem[{Samir et~al.(2024)Samir, Ahn, Prakash, Soskuthy, Shwartz, and Zhu}]{samir2024efficiently}
Farhan Samir, Emily~P Ahn, Shreya Prakash, M{\'a}rton Soskuthy, Vered Shwartz, and Jian Zhu. 2024.
\newblock Efficiently identifying low-quality language subsets in multilingual datasets: A case study on a large-scale multilingual audio dataset.
\newblock \emph{arXiv preprint arXiv:2410.04292}.

\bibitem[{Sebasti{\'a}n-Gall{\'e}s(2005)}]{sebastian2005cross}
N{\'u}ria Sebasti{\'a}n-Gall{\'e}s. 2005.
\newblock Cross-language speech perception.
\newblock \emph{The handbook of speech perception}, pages 546--566.

\bibitem[{Shan et~al.(2024)Shan, Li, Banerjee, and Oliva}]{shan2024phoneme}
Siyuan Shan, Yang Li, Amartya Banerjee, and Junier~B Oliva. 2024.
\newblock Phoneme hallucinator: One-shot voice conversion via set expansion.
\newblock In \emph{Proceedings of the AAAI Conference on Artificial Intelligence}, volume~38, pages 14910--14918.

\bibitem[{Shriberg and Lof(1991)}]{shriberg1991reliability}
Lawrence~D Shriberg and Gregory~L Lof. 1991.
\newblock Reliability studies in broad and narrow phonetic transcription.
\newblock \emph{Clinical Linguistics \& Phonetics}, 5(3):225--279.

\bibitem[{Taguchi et~al.(2023)Taguchi, Sakai, Haghani, and Chiang}]{taguchi23_interspeech}
Chihiro Taguchi, Yusuke Sakai, Parisa Haghani, and David Chiang. 2023.
\newblock \href {https://doi.org/10.21437/Interspeech.2023-2584} {Universal automatic phonetic transcription into the international phonetic alphabet}.
\newblock In \emph{INTERSPEECH 2023}, pages 2548--2552.

\bibitem[{Valk and Alum{\"a}e(2021)}]{valk2021voxlingua107}
J{\"o}rgen Valk and Tanel Alum{\"a}e. 2021.
\newblock Voxlingua107: a dataset for spoken language recognition.
\newblock In \emph{2021 IEEE Spoken Language Technology Workshop (SLT)}, pages 652--658. IEEE.

\bibitem[{Xu et~al.(2022)Xu, Baevski, and Auli}]{xu22b_interspeech}
Qiantong Xu, Alexei Baevski, and Michael Auli. 2022.
\newblock \href {https://doi.org/10.21437/Interspeech.2022-60} {Simple and effective zero-shot cross-lingual phoneme recognition}.
\newblock In \emph{Interspeech 2022}, pages 2113--2117.

\bibitem[{Yao et~al.(2023)Yao, Guo, Yang, Kang, Kuang, Yang, Jin, Lin, and Povey}]{yao2023zipformer}
Zengwei Yao, Liyong Guo, Xiaoyu Yang, Wei Kang, Fangjun Kuang, Yifan Yang, Zengrui Jin, Long Lin, and Daniel Povey. 2023.
\newblock Zipformer: A faster and better encoder for automatic speech recognition.
\newblock In \emph{The Twelfth International Conference on Learning Representations}.

\bibitem[{Yao et~al.(2025)Yao, Kang, Yang, Kuang, Guo, Zhu, Jin, Li, Lin, and Povey}]{yao2025crctc}
Zengwei Yao, Wei Kang, Xiaoyu Yang, Fangjun Kuang, Liyong Guo, Han Zhu, Zengrui Jin, Zhaoqing Li, Long Lin, and Daniel Povey. 2025.
\newblock \href {https://openreview.net/forum?id=CIs9x2ZRgh} {{CR}-{CTC}: Consistency regularization on {CTC} for improved speech recognition}.
\newblock In \emph{The Thirteenth International Conference on Learning Representations}.

\bibitem[{Yusuyin et~al.(2025)Yusuyin, Ma, Huang, Zhao, and Ou}]{yusuyin2025whistle}
Saierdaer Yusuyin, Te~Ma, Hao Huang, Wenbo Zhao, and Zhijian Ou. 2025.
\newblock Whistle: Data-efficient multilingual and crosslingual speech recognition via weakly phonetic supervision.
\newblock \emph{IEEE Transactions on Audio, Speech and Language Processing}.

\bibitem[{{\.Z}elasko et~al.(2022){\.Z}elasko, Feng, Velazquez, Abavisani, Bhati, Scharenborg, Hasegawa-Johnson, and Dehak}]{zelasko2022discovering}
Piotr {\.Z}elasko, Siyuan Feng, Laureano~Moro Velazquez, Ali Abavisani, Saurabhchand Bhati, Odette Scharenborg, Mark Hasegawa-Johnson, and Najim Dehak. 2022.
\newblock Discovering phonetic inventories with crosslingual automatic speech recognition.
\newblock \emph{Computer Speech \& Language}, 74:101358.

\bibitem[{Zhang et~al.(2021)Zhang, Zhang, Wang, Yan, Song, Huang, Li, Povey, and Wang}]{speechocean762}
Junbo Zhang, Zhiwen Zhang, Yongqing Wang, Zhiyong Yan, Qiong Song, Yukai Huang, Ke~Li, Daniel Povey, and Yujun Wang. 2021.
\newblock \href {https://doi.org/10.21437/Interspeech.2021-1259} {speechocean762: An open-source non-native english speech corpus for pronunciation assessment}.
\newblock In \emph{Interspeech 2021}, pages 3710--3714.

\bibitem[{Zhao et~al.(2018)Zhao, Sonsaat, Silpachai, Lucic, Chukharev-Hudilainen, Levis, and Gutierrez-Osuna}]{zhao2018l2}
Guanlong Zhao, Sinem Sonsaat, Alif Silpachai, Ivana Lucic, Evgeny Chukharev-Hudilainen, John Levis, and Ricardo Gutierrez-Osuna. 2018.
\newblock L2-arctic: A non-native english speech corpus.
\newblock \emph{Interspeech 2018}.

\bibitem[{Zhu et~al.(2024)Zhu, Yang, Samir, and Islam}]{zhu-etal-2024-taste}
Jian Zhu, Changbing Yang, Farhan Samir, and Jahurul Islam. 2024.
\newblock \href {https://doi.org/10.18653/v1/2024.naacl-long.43} {The taste of {IPA}: Towards open-vocabulary keyword spotting and forced alignment in any language}.
\newblock In \emph{Proceedings of the 2024 Conference of the North American Chapter of the Association for Computational Linguistics: Human Language Technologies (Volume 1: Long Papers)}, pages 750--772, Mexico City, Mexico. Association for Computational Linguistics.

\bibitem[{Zhu et~al.(2022)Zhu, Zhang, and Jurgens}]{zhu22_interspeech}
Jian Zhu, Cong Zhang, and David Jurgens. 2022.
\newblock \href {https://doi.org/10.21437/Interspeech.2022-538} {{ByT5 model for massively multilingual grapheme-to-phoneme conversion}}.
\newblock In \emph{Proc. Interspeech 2022}, pages 446--450.

\end{thebibliography}

\appendix
\newpage
\onecolumn
\section{Dataset details}
\label{app:dataset}
\subsection{Dataset Overview}

{\centering{\small \begin{longtable}{cccc}
\toprule
Language & Split & Dataset & Total Duration \\ 
\midrule
swa & train & Common Voice & 28:48:36 \\ 
 & & Fleurs & 10:06:09 \\ \midrule
spa & train & Common Voice & 404:00:29 \\ 
 & & Fleurs & 06:43:33 \\ 
 & & Multilingual Librispeech & 917:41:03 \\ \midrule
bel & train & Common Voice & 452:08:22 \\ 
 & & Fleurs & 07:18:15 \\ \midrule
tam & train & Common Voice & 76:47:49 \\ 
 & & Fleurs & 06:20:35 \\ 
 & & IISc-MILE Tamil ASR Corpus & 133:17:27 \\ \midrule
kin & train & Common Voice & 1376:18:20 \\ \midrule
eng & train & Common Voice & 1584:30:15 \\ 
 & & Librispeech & 961:03:15 \\ 
 & & Fleurs & 05:38:28 \\ \midrule
ron & train & Common Voice & 05:44:05 \\ 
 & & Fleurs & 07:38:47 \\  \midrule
ell & train & Fleurs & 07:30:24 \\  \midrule
jpn & train & Fleurs & 05:03:22 \\ 
 & & Common Voice & 09:22:26 \\  \midrule
tur & train & Fleurs & 06:25:40 \\ 
 & & Common Voice & 29:30:44 \\ \midrule
hun & train & Common Voice & 21:15:06 \\ 
 & & Fleurs & 07:00:41 \\ \midrule
mon & train & Fleurs & 08:37:49 \\ 
 & & Common Voice & 03:13:37 \\ \midrule
ind & train & Common Voice & 07:46:52 \\ 
 & & Fleurs & 06:56:13 \\ \midrule
uig & train & Common Voice & 07:36:34 \\  \midrule
ita & train & Common Voice & 83:14:54 \\ 
 & & Fleurs & 06:51:31 \\ 
 & & Multilingual Librispeech & 247:22:40 \\  \midrule
mkd & train & Fleurs & 05:08:08 \\  \midrule
urd & train & Common Voice & 04:58:30 \\ 
 & & Fleurs & 05:20:32 \\ \midrule
vie & train & Fleurs & 06:42:36 \\ 
 & & Common Voice & 01:51:31 \\ \midrule
cat & train & Common Voice & 1591:25:03 \\ 
 & & Fleurs & 05:46:13 \\  \midrule
fra & train & Common Voice & 661:14:43 \\ 
 & & Multilingual Librispeech & 1076:34:49 \\ \midrule
mya & train & Fleurs & 10:04:25 \\  \midrule
kaz & train & Kazakh Speech Dataset  & 554:47:31 \\ 
 & & Kazakh Speech Corpus & 318:25:26 \\ 
 & & Fleurs & 08:54:35 \\  \midrule
deu & train & Common Voice & 778:17:18 \\ 
 & & Multilingual Librispeech & 1966:30:30 \\ 
 & & Fleurs & 06:52:51 \\  \midrule
kir & train & Fleurs & 06:59:30 \\ \midrule
mlt & train & Fleurs & 07:29:59 \\ 
 & & Common Voice & 02:24:53 \\  \midrule
bos & train & Fleurs & 07:34:14 \\  \midrule
srp & train & Common Voice & 01:08:08 \\ 
 & & Fleurs & 08:08:18 \\  \midrule
isl & train & Fleurs & 02:06:30 \\  \midrule
ori & train & Fleurs & 02:25:20 \\ \midrule
pol & train & Fleurs & 07:13:49 \\ 
 & & Multilingual Librispeech & 103:38:57 \\ 
 & & Common Voice & 24:47:44 \\ \midrule
nld & train & Common Voice & 38:07:31 \\ 
 & & Fleurs & 05:48:46 \\ 
 & & Multilingual Librispeech & 1554:14:38 \\ \midrule
slv & train & Fleurs & 05:46:47 \\ 
 & & Common Voice & 01:24:43 \\ \midrule
tel & train & Fleurs & 05:52:07 \\ \midrule
hin & train & Common Voice & 05:17:16 \\ 
 & & Fleurs & 05:08:23 \\  \midrule
ukr & train & Fleurs & 06:41:46 \\ 
 & & Common Voice & 19:56:08 \\  \midrule 
yor & train & Common Voice & 01:20:01 \\ 
 & & Fleurs & 08:27:42 \\ \midrule
aze & train & Fleurs & 06:53:37 \\  \midrule
zho & train & Common Voice & 42:04:06 \\ \midrule
mri & train & Fleurs & 13:20:08 \\ \midrule
rus & train & Fleurs & 06:16:41 \\ 
 & & Common Voice & 37:26:56 \\ \midrule
swe & train & Common Voice & 08:10:51 \\ 
 & & Fleurs & 06:20:35 \\ \midrule
pan & train & Fleurs & 04:57:37 \\  \midrule
mar & train & Common Voice & 02:13:16 \\ 
 & & Fleurs & 09:28:59 \\ \midrule
dan & train & Fleurs & 05:45:06 \\ 
 & & Common Voice & 03:16:57 \\ \midrule
zul & train & Fleurs & 11:03:07 \\  \midrule
nob & train & Fleurs & 07:57:37 \\  \midrule
por & train & Common Voice & 22:38:41 \\ 
 & & Multilingual Librispeech & 160:57:47 \\ 
 & & Fleurs & 07:45:54 \\ \midrule
ben & train & Crowd-sourced speech for Bengali & 215:24:21 \\ 
 & & Common Voice & 31:49:44 \\ 
 & & Fleurs & 08:10:49 \\ \midrule 
bak & train & Common Voice & 139:12:22 \\  \midrule
amh & train & Fleurs & 08:15:36 \\ \midrule
est & train & Fleurs & 05:22:55 \\ 
 & & Common Voice & 05:49:26 \\ \midrule
cmn & train & Aishell-1 & 150:50:14 \\ 
 & & Fleurs & 06:02:12 \\ \midrule
ces & train & Fleurs & 06:22:34 \\ 
 & & Common Voice & 22:25:29 \\  \midrule
snd & train & Fleurs & 09:08:45 \\ \midrule
glg & train & Fleurs & 05:07:12 \\ 
 & & Common Voice & 14:01:47 \\ \midrule
uzb & train & Common Voice & 32:39:44 \\ 
 & & Fleurs & 07:35:51 \\  \midrule
nya & train & Fleurs & 08:13:52 \\  \midrule
tat & train & Common Voice & 09:29:35 \\ \midrule
kor & train & Fleurs & 05:40:36 \\ \midrule
gle & train & Fleurs & 09:18:51 \\ \midrule
eus & train & Common Voice & 15:56:07 \\  \midrule
orm & train & Fleurs & 05:06:30 \\  \midrule
mal & train & Common Voice & 00:36:24 \\ 
 & & Fleurs & 07:22:11 \\  \midrule
ara & train & Fleurs & 04:56:05 \\ 
 & & Common Voice & 31:58:14 \\ \midrule
slk & train & Common Voice & 03:26:03 \\ 
 & & Fleurs & 04:32:55 \\  \midrule
hau & train & Common Voice & 02:06:03 \\ 
 & & Fleurs & 10:05:18 \\ \midrule
yue & train & Common Voice & 03:26:30 \\ 
 & & Fleurs & 05:33:36 \\ \midrule
ceb & train & Fleurs & 09:19:35 \\ \midrule
tha & train & Fleurs & 06:12:42 \\ 
 & & Common Voice & 37:07:21 \\ \midrule
ful & train & Fleurs & 10:16:26 \\  \midrule
afr & train & Fleurs & 02:42:43 \\  \midrule
kat & train & Common Voice & 09:34:08 \\ 
 & & Fleurs & 03:52:10 \\  \midrule 
fin & train & Fleurs & 06:44:46 \\  \midrule
tgk & train & Fleurs & 06:31:01 \\ \midrule
lit & train & Fleurs & 07:16:38 \\  \midrule
sin & train & Crowd-sourced speech for Sinhala & 215:47:11 \\ \midrule
cym & train & Fleurs & 09:07:12 \\  \midrule
kmr & train & Common Voice & 04:55:01 \\ \midrule
msa & train & Fleurs & 07:17:01 \\  \midrule
jav & train & Crowd-sourced speech for Javanese & 295:46:56 \\ 
 & & Fleurs & 08:36:13 \\  \midrule
xho & train & Fleurs & 09:46:42 \\\midrule
bul & train & Fleurs & 07:02:45 \\ \midrule
ina & train & Common Voice & 04:32:09 \\ \midrule
skr & train & Common Voice & 01:17:07 \\  \midrule
hrv & train & Fleurs & 08:46:37 \\  \midrule
sna & train & Fleurs & 07:33:33 \\  \midrule
som & train & Fleurs & 09:50:14 \\ \midrule
lao & train & Fleurs & 05:34:58 \\ \bottomrule
\caption{Detailed statistics of IPAPack++. Only the train split of the original datasets were kept. Each language is represented by the ISO 639-3 standard code. }
\label{app:plus_stats}
\end{longtable}

}}

\twocolumn
The detailed breakdown of VoxAngeles can is available at \citet{chodroff-etal-2024-phonetic} and the detailed descriptions of Dorecos-IPA can be found at \citet{zhu-etal-2024-taste}. The full breakdown of individual languages is listed at Table~\ref{app:plus_stats}. 

\subsection{Final training data}
For final training data, we removed low quality samples based on the following criteria.
\begin{itemize}
    \item Audio samples longer than 24 seconds or shorter 1 second, which account for less than 0.01\% of samples.
    \item IPA sequences longer than 512 tokens or shorter than 5 tokens, as determined by the tokenizer. 
    \item IPA sequences longer than 90\% of the output frame length, which can lead to inf loss values for CTC models. The 90\% ratio also accounts for the speed perturbation. 
\end{itemize}
All data were partitioned into individual shards of 20,000 samples using the \texttt{shar} format in \texttt{lhotse}. All shards were randomly shuffled during model training. The detailed statistics can be found in Table~\ref{tab:training-stats}.

\subsection{Pseudo-labeled data}
For the VoxLingua-107 \cite{valk2021voxlingua107}, we used the original segmented sentences. For the MMS ulab V2 \cite{peng-etal-2024-owsm}, the original audios were not segmented. We also failed to apply voice activity detection due to the presence of background noises and music. So we randomly segmented the audio into individual chunks by uniformly sampling the chunk length between 1 and 20 seconds. 

Same as the original training data, all pseudo-labelled data were also partitioned into individual shards of 20,000 samples using the \texttt{shar} format in \texttt{lhotse}. The detailed statistics can be found in Table~\ref{tab:pseudo-stats}.

\begin{table}[ht]
\centering
\begin{tabular}{ll}
\toprule
\textbf{Audio count:}               & 8,289,886       \\
\textbf{Total duration (hh:mm:ss)}  & 17132:58:48 \\ 
\textbf{mean}                       & 7.4         \\
\textbf{std}                        & 4.4         \\ 
\textbf{min}                        & 1.0         \\ 
\textbf{25\%}                       & 4.2         \\ 
\textbf{50\%}                       & 5.7         \\ 
\textbf{75\%}                       & 8.7         \\ 
\textbf{99\%}                       & 19.7        \\ 
\textbf{99.5\%}                     & 20.0        \\ 
\textbf{99.9\%}                     & 20.0        \\ 
\textbf{max}                        & 24.0        \\ \bottomrule
\end{tabular}
\caption{Summary Statistics of the final labeled training data}
\label{tab:training-stats}
\end{table}

\begin{table}[ht]
\centering
\begin{tabular}{ll}
\toprule
\textbf{Audio count:}               & 4,270,280        \\
\textbf{Total duration (hh:mm:ss)}  & 11,851:31:53    \\ 
\textbf{mean}                       & 10.0           \\
\textbf{std}                        & 4.6            \\ 
\textbf{min}                        & 1.0            \\ 
\textbf{25\%}                       & 6.0            \\ 
\textbf{50\%}                       & 9.0            \\ 
\textbf{75\%}                       & 13.2           \\ 
\textbf{99\%}                       & 20.0           \\ 
\textbf{99.5\%}                     & 20.0           \\ 
\textbf{99.9\%}                     & 20.0           \\ 
\textbf{max}                        & 20.0           \\ 
\bottomrule
\end{tabular}
\caption{Summary Statistics of the pseudo-labeled training data}
\label{tab:pseudo-stats}
\end{table}

\section{Training details}
\label{app:training}

All hyperparameters for model training are presented in Table~\ref{tab:zipa-t-hyper} and~\ref{tab:zipa-cr-hyper}. Unless otherwise stated, we adopted the original hyperparameters in the Zipformer recipe in Icefall\footnote{\url{https://github.com/k2-fsa/icefall/tree/master/egs/librispeech/ASR/zipformer}}. For noisy student training, we initialized the model with the latest \textsc{Zipa-Cr} checkpoints at 500k steps for both sizes and continued to train the model by mixing the labeled data and the pseudo-labeled data at each step. 

\begin{table*}[h!]
\small
\centering
\begin{tabular}{ccc}
\toprule
\textbf{Hyperparameter}               & \textsc{Zipa-T-Small} & \textsc{Zipa-T-Large}  \\ \midrule
\textbf{Feedforward Dimensions}            & 512,768,1024,1536,1024,768                   & 768,768,1536,2048,1536,768                                     \\ 
\textbf{Encoder Dimensions} & 192,256,384,512,384,256 & 512,512,768,1024,768,512\\
\textbf{Num. Layers}                  & 2,2,3,4,3,2                     & 4,3,4,5,4,4                                         \\ 
\textbf{Downsampling factors}         & \multicolumn{2}{c}{1,2,4,8,4,2}                            \\ 
\textbf{Output downsampling factor}         & \multicolumn{2}{c}{4}                            \\ 
\textbf{Joiner dimension} & 512 & 1024 \\
\textbf{Decoder dimension} & 512 & 1024\\
\textbf{Parameters}                   & 65M                   & 302M                               \\ 
\textbf{Initial Learning Rate}        &  0.035                  & 0.025                               \\
\textbf{Optimizer} & \multicolumn{2}{c}{Scaled Adam}             \\ 
\textbf{Scheduler}                    & \multicolumn{2}{c}{Eden Scheduler}             \\ 
\textbf{Total Training Steps}         & \multicolumn{2}{c}{500k}          \\ 
\textbf{Effective Batch Size}        & 800 seconds                    & 600 seconds                                      \\ 
\textbf{Mixed Precision}              & \multicolumn{2}{c}{bfloat16}                            \\ 
\textbf{GPUs} & A40 48G & 2 $\times$ A100 40G \\
\textbf{Training time} & 5 days & 4 days \\\bottomrule
\end{tabular}
\caption{Hyperparameters for \textsc{Zipa-T} models.}
\label{tab:zipa-t-hyper}
\end{table*}

\begin{table*}[h!]
\small
\centering
\begin{tabular}{ccc}
\toprule
\textbf{Hyperparameter}               & \textsc{Zipa-Cr-Small} & \textsc{Zipa-Cr-Large}  \\ \midrule
\textbf{Feedforward Dimensions}            & 512,768,1024,1536,1024,768                   & 768,768,1536,2048,1536,768                                     \\ 
\textbf{Encoder Dimensions} & 192,256,384,512,384,256 & 512,512,768,1024,768,512\\
\textbf{Num. Layers}                  & 2,2,3,4,3,2                     & 4,3,4,5,4,4                                         \\ 
\textbf{Downsampling factors}         & \multicolumn{2}{c}{1,2,4,8,4,2}                            \\ 
\textbf{Output downsampling factor}         & \multicolumn{2}{c}{2}                            \\ 
\textbf{Parameters}                   & 64M                   & 300M                               \\ 
\textbf{Initial Learning Rate}        & 0.035                  & 0.025                                  \\
\textbf{SpecAug: Num. frame masks} &\multicolumn{2}{c}{20} \\
\textbf{SpecAug: Max mask fraction} & \multicolumn{2}{c}{0.3} \\
\textbf{Optimizer} & \multicolumn{2}{c}{Scaled Adam}             \\ 
\textbf{Scheduler}                    & \multicolumn{2}{c}{Eden Scheduler}             \\ 
\textbf{Total Training Steps}         & \multicolumn{2}{c}{500k}          \\ 
\textbf{Effective Batch Size}        & 500 seconds                    & 240 seconds                                      \\ 
\textbf{Mixed Precision}              & \multicolumn{2}{c}{bfloat16}                            \\ 
\textbf{GPUs} & A40 48G & 2 $\times$ A100 40G \\
\textbf{Training time} & 6 days & 4 days \\\midrule
\textbf{Noisy student training} & \textsc{Zipa-Cr-Ns-Small} & \textsc{Zipa-Cr-Ns-Large}\\\midrule
\textbf{Steps} & 200k & 280k \\
\textbf{Training time} & 5 days & 4 days \\
\textbf{Initial learning rate} & \multicolumn{2}{c}{1e-3} \\    
\bottomrule
\end{tabular}
\caption{Hyperparameters for \textsc{Zipa-T} models.}
\label{tab:zipa-cr-hyper}
\end{table*}


\end{document}